\documentclass[11pt]{article}

  \usepackage[final]{acl}

 \usepackage{times}
\usepackage{latexsym}

 \usepackage[T1]{fontenc}
   
 \usepackage[utf8]{inputenc}

   \usepackage{microtype}

   \usepackage{inconsolata}

  \usepackage{graphicx}

\usepackage{booktabs}
\usepackage{tabularx}

 \usepackage{algorithm}
\usepackage{algpseudocode}
 \usepackage{enumitem}
\usepackage{hyperref}
\usepackage{listings}
\usepackage{makecell}
\usepackage{multirow}
\usepackage{soul}
  
 \providecommand{\mydataset}{{BIRD-History}}

\definecolor{codebgcolor}{gray}{0.95}
\definecolor{picolor}{RGB}{36,175,255}
\definecolor{krcolor}{RGB}{0,106,183}
\definecolor{socolor}{RGB}{255,144,116}
\definecolor{kicolor}{RGB}{203,31,3}

\setstcolor{red}
  \providecommand{\DIFadd}[1]{{\protect\color{black}#1}}
\providecommand{\DIFdel}[1]{}

 \title{\mydataset: A Benchmark for History-Driven Text-to-SQL with Fine-Grained Knowledge Annotations}

\author{
\normalsize
 Yunfan Zhou\textsuperscript{1}, Qiming Shi\textsuperscript{1}, Yizhou Yang\textsuperscript{2}, Di Weng\textsuperscript{2,$\dagger$}, Yingcai Wu\textsuperscript{1} 
\\ 
\textsuperscript{1}State Key Lab of CAD\&CG, Zhejiang University\\
\textsuperscript{2}School of Software Technology, Zhejiang University\\
\textsuperscript{$\dagger$}Corresponding author \\
\texttt{\{yf.zhou,qimingshi,yangyizhou,dweng,ycwu\}@zju.edu.cn}
}

\begin{document}
\maketitle
\begin{abstract}
While recent Large Language Model (LLM)-based text-to-SQL systems achieve impressive performance on standard benchmarks,
they struggle when user queries implicitly rely on domain-specific knowledge, such as business logic, data conventions, and analytical practices, that is neither captured by the schema nor explicitly stated in the natural language question.
Historical SQL query logs offer a valuable source of such knowledge, yet existing benchmarks do not adequately support evaluation of history-driven approaches.
 To address this gap, we introduce BIRD-History, a benchmark consisting of 1,393 tasks across 11 databases, designed to evaluate text-to-SQL systems' ability to ground underspecified natural language questions using historical SQL scripts.
Each task is annotated with ground-truth labels specifying which historical queries contain relevant knowledge and which SQL clauses encode it, enabling systematic evaluation of both retrieval effectiveness and knowledge utilization. 
 \DIFdel{Alongside the benchmark, we propose a plug-in retriever that extracts five types of external knowledge from historical SQL scripts, then retrieves and reranks relevant fragments for query generation.
The retriever integrates seamlessly into existing few-shot text-to-SQL pipelines without requiring prompt modifications. }
\DIFadd{We further introduce a plug-in retriever that extracts relevant knowledge from historical logs, seamlessly enhancing existing few-shot text-to-SQL pipelines without architectural modifications.}
Experiments demonstrate consistent improvements across four text-to-SQL systems, highlighting the value of leveraging historical query logs for handling underspecified queries.
    Dataset and code are open-sourced on \url{https://github.com/zjuidg/BIRD-History}.
\end{abstract}

\section{Introduction}
Text-to-SQL, which converts natural language questions into database queries, has emerged as a powerful tool for democratizing data exploration and analysis~\cite{nl2sql-survey, next-gen-nl2sql-survey, deep-learning-nl2sql-survey}.
Recent text-to-SQL approaches~\cite{alphasql, opensearch-sql, dpo-text2sql, mac-sql, din-sql} utilizing Large Language Models (LLMs) have driven remarkable progress, with state-of-the-art systems achieving over 90\% accuracy on the Spider~\cite{spider} benchmark and 80\% on the more challenging BIRD~\cite{bird} benchmark.

\begin{figure}[t]
  \includegraphics[width=\columnwidth]{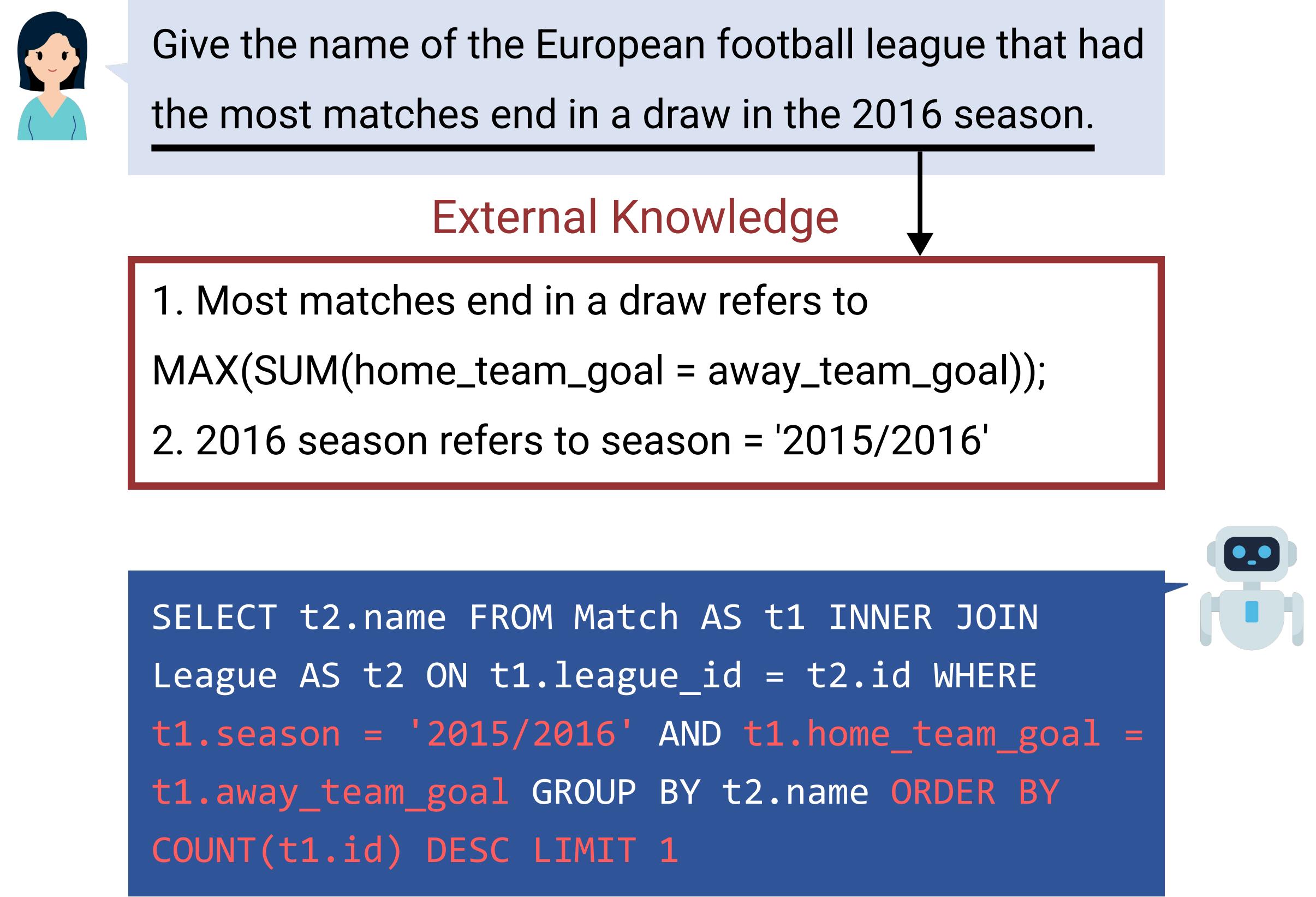}
  \caption{A text-to-SQL example from the BIRD dataset involving external knowledge.}
  \label{fig:motivated-examples}
\end{figure}

Current text-to-SQL approaches primarily rely on schema linking to match database elements with natural language tokens based on few-shot learning~\cite{din-sql} or dynamic retrieval~\cite{opensearch-sql, dail-sql, codes}.
However, while these techniques effectively learn common query patterns across domains, they often struggle when user queries implicitly rely on nuanced domain-specific knowledge, such as business logic, data conventions, and analytical practices, that is neither captured by the schema nor explicitly stated in the natural language question.
For instance, in a European football match database, asking for the league with ``the most matches end in a draw'' requires understanding that draws are identified by comparing home and away team goals, and that ``2016 season'' corresponds to the value ``2015/2016'' in the database (\autoref{fig:motivated-examples}).
Such domain-specific knowledge is hard to infer from the schema alone and varies significantly across different domains.
\DIFadd{While explicit knowledge bases offer one solution to capturing domain-specific information, they require substantial manual effort from domain experts to construct and maintain, particularly as database schemas and business logic evolve over time.}

Recent work has recognized historical SQL queries as a valuable knowledge source for text-to-SQL generation. 
TailorSQL~\cite{tailorsql} shows that past query workloads implicitly encode domain-specific information, such as common join paths and column semantics, that is not apparent from the schema alone. 
However, existing benchmarks do not adequately support evaluation of such history-driven approaches, as they either lack historical query contexts or do not provide ground-truth labels linking queries to the specific historical SQL fragments required for their generation.

To address this gap, we introduce \mydataset, a benchmark for evaluating text-to-SQL systems' ability to ground underspecified natural language queries using historical SQL scripts.
\mydataset\ is built upon the BIRD dataset~\cite{bird} and consists of 1,393 tasks across 11 databases.
The construction of \mydataset\ follows a hybrid approach combining LLM-based generation under carefully designed constraints, LLM-as-a-judge refinement, and human expert cross-validation.
For each task, a text-to-SQL system is provided with three types of inputs:
the development set of the original BIRD dataset as historical SQL queries, database schemas, and an underspecified natural language question.
Crucially, each task is annotated with ground-truth labels specifying which historical queries contain knowledge relevant to the target query, along with the specific SQL clauses (e.g., \texttt{WHERE}, \texttt{LIMIT}, \texttt{JOIN}) that encode this knowledge.
These fine-grained annotations enable systematic evaluation of both retrieval effectiveness (i.e., how well systems identify relevant historical queries) and knowledge utilization (i.e., whether systems extract and apply the correct SQL fragments for generation).

Based on \mydataset, we further propose a plug-in retriever that extracts fine-grained knowledge fragments from historical SQL queries to improve text-to-SQL generation.
The retriever operates in two stages: offline and online.
In the offline stage, it parses historical SQL queries into code fragments corresponding to five types of external knowledge, including calculation, condition, relation, dimension, and output.
These code fragments are then translated into natural language forms using LLMs to facilitate effective retrieval.
In the online stage, given a user question, the retriever first broadly recalls relevant knowledge based on semantic similarity.
The recalled knowledge is subsequently refined through LLM-based reranking and filtering to ensure precision.
The retrieval results can be seamlessly integrated into existing dynamic few-shot text-to-SQL pipelines without requiring modifications to their prompt templates.
Empirical evaluation on \mydataset\ demonstrates consistent improvements across four existing text-to-SQL systems.

In summary, we make three key contributions:

\begin{itemize}[itemsep=0pt, parsep=0pt]
\item We present \mydataset, a benchmark specifically designed to evaluate text-to-SQL systems' capability to ground external knowledge from historical query logs.
\item We introduce a plug-in retriever that extracts and retrieves fine-grained external knowledge from historical SQL queries.
\item We demonstrate through experiments that our approach outperforms prior retrieval methods for text-to-SQL generation.
\end{itemize}

\section{Related Work}
\textbf{Text-to-SQL approaches}.
\DIFadd{Early text-to-SQL approaches used structured decoding~\cite{coarse2fine, smbop} and schema-aware encoding~\cite{represent-schema-structure, rat-sql, slsql}.
Recent LLM-based approaches leverage in-context learning and task decomposition~\cite{din-sql, route, dcg-sql, share, omnisql}.
To improve accuracy, preprocessing methods in text-to-SQL pipelines focus on schema linking~\cite{resdsql, mac-sql, chess}, database value retrieval~\cite{irnet, bridge, tabert}, and external knowledge acquisition~\cite{dail-sql, codes, regroup}.}

\DIFadd{External knowledge has gained increasing attention, with benchmarks like KaggleDBQA~\cite{kaggledbqa}, Spider-DK~\cite{spider-dk}, BIRD~\cite{bird}, and Spider 2.0~\cite{spider2} incorporating it as input.
Early work like DIN-SQL~\cite{din-sql} used static few-shot examples.
Recent approaches shifted to dynamic few-shot selection:
CodeS~\cite{codes} and AID-SQL~\cite{aid-sql} use ranking models for semantic matching, while DAIL-SQL~\cite{dail-sql}, MCS-SQL~\cite{mcs-sql}, and OpenSearch-SQL~\cite{opensearch-sql} combine question similarity with SQL skeleton matching.}

\DIFdel{Text-to-SQL translation is the task of automatically converting natural language questions into executable SQL queries over relational databases.
Early work established foundational techniques through structured decoding approaches~\mbox{\cite{coarse2fine, smbop}} and schema-aware encoding methods~\mbox{\cite{represent-schema-structure, rat-sql, slsql}}.
With the advent of LLMs, recent approaches have focused on leveraging in-context learning, task decomposition, and data synthesis techniques to enhance cross-domain generalization and query generation accuracy~\mbox{\cite{din-sql, route, dcg-sql, share, omnisql}}.
To improve translation accuracy, recent preprocessing methods of text-to-SQL focus on acquiring relevant context through schema linking, database value retrieval, and external knowledge acquisition~\mbox{\cite{nl2sql-survey}}.
Schema linking identifies relevant tables and columns for the natural language query~\mbox{\cite{resdsql, mac-sql, chess}}
Database value retrieval extracts relevant cell values from database tables according to natural language mentions~\mbox{\cite{irnet, bridge, tabert}}.
External knowledge acquisition enhances text-to-SQL by incorporating domain-specific information through external resources, enabling models to handle specialized terminology and complex calculations that require knowledge beyond what is available in the database schema alone~\mbox{\cite{dail-sql, codes, regroup}}.}

\DIFdel{Among the abovementioned contextual information, external knowledge has gained increasing attention in recent text-to-SQL research, with several benchmarks such as KaggleDBQA~\mbox{\cite{kaggledbqa}}, Spider-DK~\mbox{\cite{spider-dk}}, BIRD~\mbox{\cite{bird}}, and Spider 2.0~\mbox{\cite{spider2}} incorporating it as an integral part of the input.
Early work like DIN-SQL~\mbox{\cite{din-sql}} employed static few-shot examples, providing fixed demonstrations to guide query generation.
More recent approaches have shifted toward dynamic few-shot selection, which retrieves task-specific examples based on the input query.
CodeS~\mbox{\cite{codes}} and AID-SQL~\mbox{\cite{aid-sql}} advance retrieval through ranking models to better capture the semantic correspondence between natural language questions and SQL queries.
DAIL-SQL~\mbox{\cite{dail-sql}}, MCS-SQL~\mbox{\cite{mcs-sql}}, and OpenSearch-SQL~\mbox{\cite{opensearch-sql}} combine question similarity with SQL skeleton matching to select relevant demonstrations.
These dynamic approaches have demonstrated superior performance by adapting the provided context to each specific query.}

\textbf{Text-to-SQL datasets}.
\DIFadd{Classic benchmarks like Spider~\cite{spider} focus on cross-domain generalization.
Extensions incorporate external knowledge~\cite{kaggledbqa, bird, spider2} or evaluate robustness~\cite{spider-syn, dr-spider, spider-dk}.
Recent work includes PARROT~\cite{parrot} for cross-system SQL translation and SWE-SQL~\cite{swe-sql} for real-world SQL issue resolution.}

\DIFadd{However, existing benchmarks provide external knowledge as natural language documentation~\cite{kaggledbqa, spider-dk, bird, spider2}, whereas real-world query logs often lack annotations.
Current dynamic few-shot approaches retrieve syntactically similar examples from cross-domain datasets, which cannot capture dataset-specific conventions~\cite{tailorsql}.
We introduce \mydataset, a benchmark for evaluating text-to-SQL systems' ability to leverage historical query logs, with a plug-in baseline retriever for existing few-shot pipelines.}

\DIFdel{Numerous benchmarks have been proposed to evaluate text-to-SQL systems under different settings.
Classic cross-domain benchmarks such as Spider~\mbox{\cite{spider}} focus on generalization across diverse database schemas but often contain questions with explicit schema mentions.
Several benchmarks extend this foundation by incorporating external knowledge~\mbox{\cite{kaggledbqa, bird, spider2}} or evaluating robustness against various perturbations~\mbox{\cite{spider-syn, dr-spider, spider-dk}}.
For instance, Spider-Syn~\mbox{\cite{spider-syn}} replaces schema-related words with synonyms to test model robustness to paraphrasing.
More recently, PARROT~\mbox{\cite{parrot}} evaluates cross-system SQL translation across different database dialects,
and SWE-SQL~\mbox{\cite{swe-sql}} focuses on resolving user-reported SQL issues in real-world applications.}

\DIFdel{However, existing benchmarks predominantly provide external knowledge in the form of natural language documentation~\mbox{\cite{kaggledbqa, spider-dk, bird, spider2}}, whereas in real-world scenarios, past user queries often lack sufficient annotations, presenting additional challenges for knowledge extraction.
Furthermore, current dynamic few-shot approaches in text-to-SQL retrieve syntactically similar examples from cross-domain training datasets, which cannot adequately capture dataset-specific conventions and domain assumptions unique to each database~\mbox{\cite{tailorsql}}.
Motivated by these observations, we introduce \mbox{\mydataset}, a benchmark for evaluating text-to-SQL systems' ability to leverage historical query logs, along with a plug-in retriever that can be integrated into existing few-shot pipelines to extract and utilize domain knowledge from past queries.}

\section{\mydataset\ Dataset}
\subsection{Task Definition}
\label{ssec:task-def}

Following established text-to-SQL frameworks~\cite{bird, spider2}, we formalize the task of history-grounded text-to-SQL as follows:
Given a natural language question $Q$,
a database $\mathcal{D}$ containing schemas and values,
and a collection of historical SQL queries $\mathcal{H} = \{S_1, S_2, ..., S_n\}$ \DIFdel{previously}\DIFadd{successfully} executed on $\mathcal{D}$,
the goal is to generate an executable and correct SQL query $Y$.

The key challenge lies in effectively leveraging the historical query collection $\mathcal{H}$.
Since $\mathcal{H}$ can be large and contain queries irrelevant to the current question,
directly incorporating all historical queries into the model context incurs significant computational costs and introduces irrelevant noise that may degrade performance.
Therefore, we introduce an intermediate retrieval step that selects a relevant subset $\mathcal{H}' \subset \mathcal{H}$ based on the question $Q$ and schema $\mathcal{D}$.
Following the general Retrieval-Augmented Generation (RAG) paradigm~\cite{rag-llm-survey}, we decompose the task into two stages:

\textbf{Retrieval Stage:} A retriever $R$ selects relevant historical queries:
\begin{equation}
\mathcal{H}' = R(Q, \mathcal{D}, \mathcal{H} \mid \theta_R)
\end{equation}
where $\theta_R$ represents the retriever's parameters.

\textbf{Generation Stage:} A text-to-SQL model $f$ generates the target SQL query:
\begin{equation}
Y = f(Q, \mathcal{D}, \mathcal{H}' \mid \theta)
\end{equation}
where $\theta$ denotes the model parameters.

\subsection{Data Source}

We construct \mydataset\ by treating the SQL scripts in the BIRD development set as historical queries $\mathcal{H}$ and generating new evaluation tasks on the same databases.
We choose BIRD because each database contains a substantial number of SQL queries (139 queries per database in the development set on average), providing sufficient historical context for knowledge extraction and retrieval evaluation.
Additionally, BIRD features large-scale schemas and cross-domain coverage, enabling evaluation across diverse scenarios.

\subsection{Dataset Construction}

\begin{figure}[ht]
  \includegraphics[width=\columnwidth]{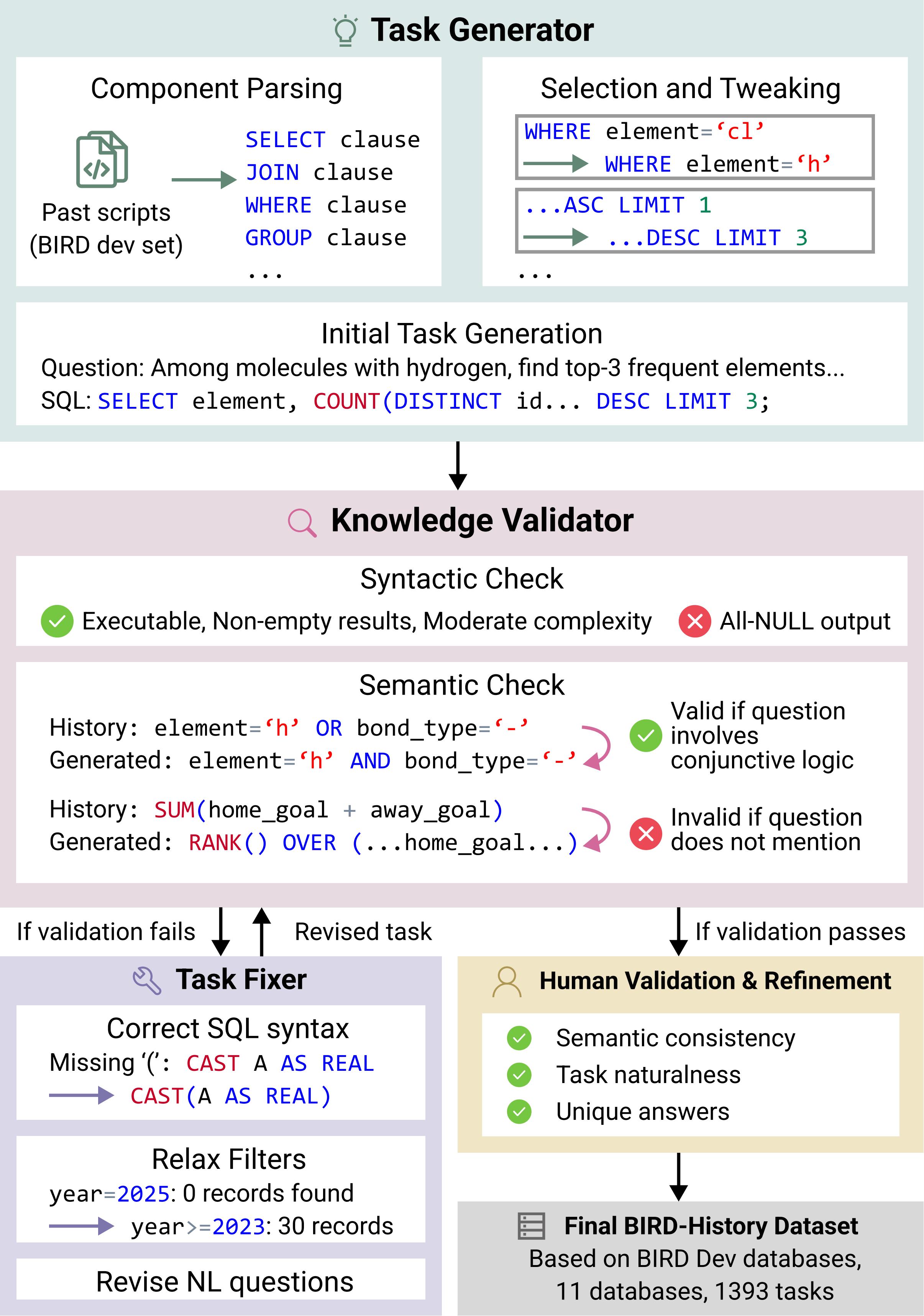}
  \caption{The \mydataset\ dataset construction pipeline. The process consists of two stages: (1) Automatic task generation, where a Task Generator parses historical SQL scripts into reusable components and creates new question-SQL pairs, a Knowledge Validator performs syntactic and semantic checks to ensure quality and proper knowledge reuse, and a Task Fixer iteratively refines candidates that fail validation; (2) Human validation and refinement, where domain experts cross-validate semantic consistency, task naturalness, and answer uniqueness to produce the final dataset.}
  \label{fig:dataset-construction}
\end{figure}

We employ a two-stage semi-automatic construction process that balances construction efficiency and data quality, as shown in \autoref{fig:dataset-construction}.

\subsubsection{Stage 1: Automatic Task Generation}
\label{sssec:auto-task-gen}

\textbf{Task Generator}.
\DIFdel{The generator creates initial question-SQL pairs by identifying and reusing SQL components from historical queries.}
For each task, it randomly samples SQL scripts from the same database in the BIRD development set and parses each script into clause-level components using SQLGlot\footnote{\url{https://sqlglot.com/sqlglot.html}}.
The underlying LLM then selects logically related components \DIFadd{(e.g., derivation of two attributes \texttt{SELECT...AS A} and \texttt{SELECT...AS B} from two queries)} which serve as building blocks for a more complex query.
\DIFadd{The generator is instructed to synthesize new logic (e.g., \texttt{SUM(A)/(SUM(A)+SUM(B))}, which does not exist in the history) rather than simply concatenating existing fragments. Additionally,}\DIFdel{To simulate realistic query variations}
it randomly tweaks literal values (e.g., values in the \texttt{WHERE} clause)
or basic operators (e.g., replacing \texttt{>} with \texttt{<}) before generating the SQL task \DIFadd{to simulate realistic query variations}.
Table and column names of the selected components are kept unchanged to preserve the underlying knowledge patterns.
For natural language question generation, the generator provides example BIRD questions to guide the LLM in producing queries that match the dataset's natural language style \DIFadd{(prompt in \autoref{lst:task_generator})}.

\textbf{Knowledge Validator}.
The validator performs multi-faceted verification to ensure generated tasks meet quality standards and appropriately reuse the historical knowledge \DIFadd{(\autoref{lst:knowledge_validator})}.
It first checks basic executability criteria by verifying that the SQL is syntactically valid, produces non-empty results, and avoids trivial outputs such as all-NULL or all-zero values.
It then examines the task's structural complexity, ensuring moderate difficulty (e.g., approximately 20-30 lines of code) without excessively complex joins that would result in impractical execution times.
Beyond these basic checks, the validator performs semantic verification to detect invalid combinations of knowledge and new computational logic.
For instance, if the historical queries filter transactions using conjunctive conditions \texttt{WHERE transaction\_type='deposit' AND status='completed'}, 
the generated SQL should not arbitrarily change to disjunctive logic unless the natural language question explicitly requested.
Similarly, if no historical query performs ranking calculations, 
the generated SQL should not introduce window functions unless the question explicitly mentions ranking or percentile operations.
If any issues are detected, the validator returns the diagnostic feedback to the task fixer for regeneration.

\textbf{Task Fixer}.
The fixer addresses issues identified by the validator through targeted revisions \DIFadd{(\autoref{lst:task_fixer})}.
For instance, it corrects SQL syntax if execution errors are detected.
For empty result sets, it relaxes filtering conditions,
while for unreasonable results (e.g., all-zero values), it revises both the question and SQL.
The fixer also optimizes queries by removing redundant joins and subqueries.
Throughout the revision process, the fixer adheres to the rules of knowledge reuse defined by the validator.
This validation-fixing loop allows up to three revision attempts before rejecting the candidate task, ensuring consistency with the scope of knowledge in the original BIRD dataset while maintaining data quality.

\subsubsection{Stage 2: Human Validation and Refinement}

\DIFdel{Following automatic generation, three authors with SQL expertise conduct cross-validation of all candidate tasks to ensure dataset quality.
First, they verify the semantic consistency between natural language questions and the corresponding SQL queries.
Second, they assess the naturalness of generated tasks, filtering out those that lack coherent query purpose.
This ensures that \mbox{\mydataset} includes new, more complex query tasks that appropriately leverage historical knowledge rather than simply stacking existing queries.
Third, they remove tasks with non-unique answers.}
\DIFadd{In order to ensure dataset quality, three authors with SQL expertise act as annotators to conduct cross-validation of all candidate tasks.
They verify semantic consistency between natural language questions and SQL queries, and check task naturalness.
Moreover, they remove tasks with non-unique answers.}
For example, a question asking to ``find football teams with many goals'' could refer to different thresholds such as more than the average or more than ten goals.
Unlike underspecified queries that can be resolved by leveraging external knowledge, these ambiguous questions lack a deterministic answer even with complete historical context, as the interpretation of vague terms like ``many'' varies across different query sessions.
\DIFdel{Any issues identified during validation are manually corrected by the authors through discussion until consensus is reached.}

\DIFadd{To assess annotation reliability, we measure inter-annotator agreement on the validation decisions.
The three annotators achieve a full agreement rate of 0.80. 
Since the labels are highly imbalanced (most candidates are valid), we report Gwet’s AC1 instead of $\kappa$ statistics, obtaining AC1 = 0.82, indicating strong agreement.
For cases without full agreement, annotators first conduct a discussion to revise the annotation.
If consensus is not reached, the final decision is made by a senior expert with over 10 years of SQL authoring experience.}

\subsection{Dataset Statistics}

\begin{table*}[ht]
\centering
\small
\setlength{\tabcolsep}{1pt}
\begin{tabularx}{\textwidth}{l*{8}{>{\centering\arraybackslash}X}}
\toprule
\textbf{Dataset} & \textbf{\# Tasks} & \textbf{\# Tok. / SQL} & \textbf{\# Func. / SQL} & \textbf{\# Subq. / SQL} & \textbf{JOIN Len. / SQL} & \textbf{Nest Depth / SQL} & \textbf{External Knowledge} & \textbf{Retrieval Annotations} \\
\midrule
WikiSQL & 15,878 & 25.92 & 0.29 & 0 & 1 & 1 & No & No \\
Spider & 2,147 & 30.73 & 0.57 & 0.08 & 1.53 & 1.08 & No & No \\
KaggleDBQA & 185 & 27.37 & 0.73 & 0.03 & 1.21 & 1.02 & \textbf{Yes} & No \\
SEDE & 857 & 93.18 & 2.97 & 0.31 & 1.76 & 1.25 & No & No \\
BIRD (dev) & 1,534 & 53.38 & 1.74 & 0.10 & 1.92 & 1.09 & \textbf{Yes} & No \\
Spider 2.0-lite & 256 & 388.85 & 7.27 & 2.63 & 2.11 & 1.96 & \textbf{Yes} & No \\
Spider 2.0-snow & 120 & \textbf{527.90} & \textbf{10.33} & \textbf{2.83} & 2.16 & 1.83 & \textbf{Yes} & No \\
\midrule
\mydataset & 1,393 & 181.70 & 5.87 & 1.98 & \textbf{2.57} & \textbf{2.06} & \textbf{Yes} & \textbf{Yes} \\
\bottomrule
\end{tabularx}
\caption{\DIFadd{Statistical comparison among \mydataset\ and other text-to-SQL benchmarks, where Tok., Func., Subq., JOIN Len., and Nest Depth denote the average number of tokens, functions, subqueries, JOIN length, and nesting depth per SQL query, respectively.}}
\label{tab:comparison}
\end{table*}

\mydataset\ comprises 1,393 evaluation tasks distributed across all 11 databases in BIRD development set.
On average, each task requires reusing 5.3 historical SQL fragments, with the number of reused fragments ranging from 1 to 11.
\DIFadd{As shown in \autoref{tab:comparison},} the generated SQL queries contain an average of 181.70 tokens, approximately 3.4 times longer than BIRD queries (53.38 tokens on average), which corresponds to roughly 30 lines of formatted code\footnote{Line count is measured using \href{https://sqlglot.com/sqlglot.html}{SQLGlot} formatting with default indentation settings.}.
\DIFadd{See Appendix~\ref{appssec:dataset-analysis} for detailed dataset statistics and analysis.}

\section{Baseline Approach}
\label{sec:baseline-approach}
\begin{figure}[t]
  \includegraphics[width=\columnwidth]{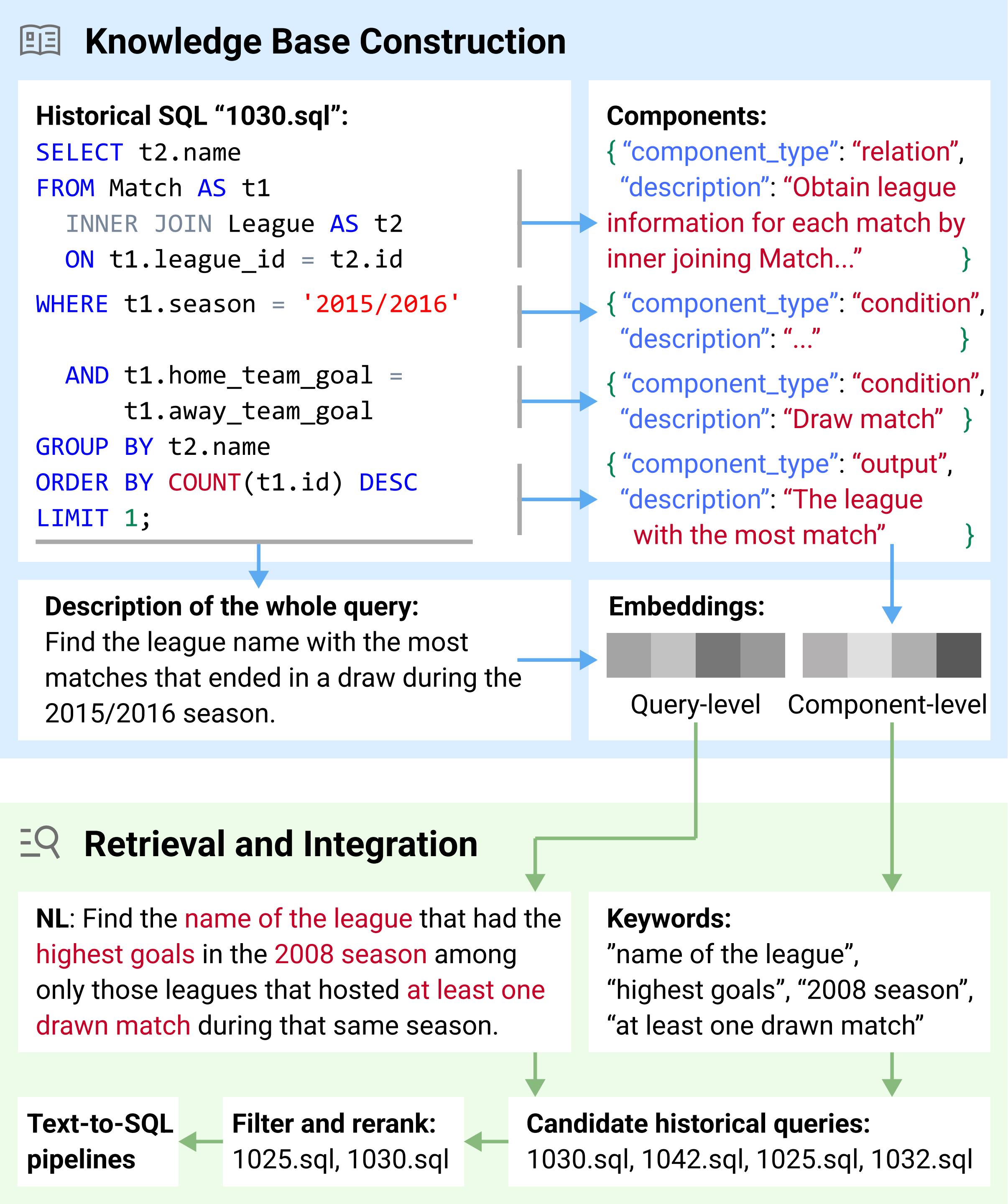}
  \caption{The workflow of our baseline retriever. Historical SQL queries are decomposed into components with natural language descriptions and embeddings (top). For each natural language query, the retriever extracts keywords, retrieves candidates via semantic similarity at both query and component levels\DIFdel{, then filters and reranks using an LLM} before integration into text-to-SQL pipelines (bottom).}
  \label{fig:system-workflow}
\end{figure}

To effectively leverage historical SQL queries for improved text-to-SQL generation, we propose a basic retriever (\autoref{fig:system-workflow}) that can be integrated as a plug-in to existing text-to-SQL pipelines.

\DIFadd{\textbf{External Knowledge Categorization}.
We categorize knowledge in historical SQL into five types based on SQL grammar (\autoref{tab:knowledge_type} in Appendix~\ref{appssec:knowledge_type_cate}): Calculation (formulas in \texttt{SELECT}/\texttt{ORDER BY}),
Condition (predicates in \texttt{WHERE}/\texttt{HAVING}),
Relation (join patterns in \texttt{FROM}/\texttt{JOIN}),
Dimension (aggregation in \texttt{GROUP BY}),
and Output (result specifications in \texttt{ORDER BY}/\texttt{LIMIT}/\texttt{DISTINCT}).}

\DIFdel{\textbf{External knowledge categorization}.
We categorize external knowledge embedded in historical SQL queries into five types based on their syntactic positions in SQL grammar, following SQLite specifications\footnote{\url{https://www.sqlite.org/index.html}} (\mbox{\autoref{tab:knowledge_type}}).
\textit{Calculation} refers to formulas and functions in \texttt{SELECT} and \texttt{ORDER BY} clauses.
\textit{Condition} captures filtering predicates in \texttt{WHERE} and \texttt{HAVING} clauses.
\textit{Relation} specifies table join patterns in \texttt{FROM} and \texttt{JOIN} clauses.
\textit{Dimension} determines aggregation granularity specified in \texttt{GROUP BY} clauses.
\textit{Output} defines result presentation preferences in \texttt{ORDER BY}, \texttt{LIMIT}, and \texttt{DISTINCT} clauses, such as returning top-k results or ensuring uniqueness of returned records.}

\DIFadd{\textbf{Knowledge Base Construction}.
As shown in \autoref{fig:system-workflow}, the retriever analyzes historical queries using Abstract Syntax Tree (AST) parsing to extract structured components.
For calculation and condition knowledge, it extracts expressions by splitting at logical operators.
For relation, dimension, and output knowledge, it extracts complete clauses.
Each component is converted to natural language descriptions via LLM prompts (\autoref{lst:knowledge_extraction} in Appendix~\ref{appsssec:know_ext_prompt}), producing both query-level summaries and component-level explanations.
Dense embeddings are computed using a pre-trained encoder, with associations maintained to source queries.}

\DIFdel{\textbf{Knowledge base construction}.
As is shown in \mbox{\autoref{fig:system-workflow}}, the retriever parses historical SQL queries into structured components corresponding to the five types of knowledge using Abstract Syntax Tree (AST) parsing.
For calculation and condition knowledge, it extracts expressions by splitting at commas and logical operators while preserving nested structures.
For relation, dimension, and output knowledge, it extracts clauses as complete units.
To enable semantic matching, the retriever generates natural language descriptions for each component using an LLM prompted with database schema and column descriptions.
This produces both query-level descriptions summarizing overall intent and component-level descriptions explaining detailed external knowledge.
The retriever computes dense vector representations for all descriptions using a pre-trained sentence encoder.
To support downstream SQL generation, it maintains associations between components and their source SQL queries, enabling retrieval of complete historical queries alongside relevant fragments.}

\DIFadd{\textbf{Retrieval and Integration}.
Given an NL question, the retriever performs dense retrieval over query-level embeddings to identify candidate historical queries.
It additionally decomposes the NL question into phrases (\autoref{lst:keyword_extraction}) and retrieves relevant components by matching against component-level embeddings.
Retrieved components are traced back to their source queries and merged with initially retrieved queries.
An LLM then filters and reranks candidates based on relevance (\autoref{lst:filter-rarank}), providing complete SQL scripts and parsed components with natural language descriptions.
Results are formatted according to the target text-to-SQL pipeline's prompt template and prepended without architectural modifications.}

\DIFdel{\textbf{Retrieval and integration}.
Given an input natural language question, the retriever employs a two-stage process (\mbox{\autoref{fig:system-workflow}}).
In the \textit{retrieval} stage, it retrieves complete historical queries by computing similarity between the question embedding and query-level description embeddings stored in knowledge base.
It then decomposes the question into phrases and retrieves relevant components by matching these phrases against component-level description embeddings.
The retriever traces these components back to their source SQL queries and merges them with the initially retrieved queries to form a candidate set.
In the \textit{refinement} stage, an LLM filters and re-ranks the candidate set based on relevance to the input question.
For each candidate query, the retriever provides the complete SQL script and its parsed components, with corresponding natural language descriptions generated during knowledge base construction.
This additional context enables the LLM to better assess whether the external knowledge embedded in these components is relevant to the user's question.
The final set of retrieved queries is formatted according to the target text-to-SQL pipeline's prompt template and prepended to the input, without modifications to the pipeline's internal architecture or generation mechanism.}

\section{Experiments}
\DIFdel{We conduct experiments on our \mbox{\mydataset} benchmark to evaluate the effectiveness of our proposed retriever when integrated into existing text-to-SQL pipelines.}
\DIFadd{We conduct experiments on \mydataset, which uniquely requires retrieval from historical queries rather than providing gold context directly (as in BIRD/Spider 2.0).
This enables joint evaluation of retrieval quality and generation accuracy.}

\subsection{Settings}
\label{ssec:settings}

\textbf{Pipelines}.
We select four representative LLM-based, few-shot text-to-SQL systems:
DAIL-SQL~\cite{dail-sql}, OpenSearch-SQL~\cite{opensearch-sql}, N-rep~\cite{n-rep}, and CodeS~\cite{codes}.
Since DAIL-SQL provides two retrieval approaches, in \autoref{tab:results-in-dataset} and \autoref{tab:retriever-performance} we denoted them as ``MQS'', which selects examples based solely on masked question similarity,
and ``DAIL'', which first generates a preliminary SQL query, then ranks candidates by masked question similarity and filters them using SQL skeleton similarity thresholds.\DIFdel{Like DAIL-SQL, OpenSearch-SQL first applies MQS to select similar queries by masking database-specific tokens, then enhances the retrieved Query-SQL pairs with chain-of-thought.}
N-rep builds upon the MQS approach, but additionally filters the retrieval pool to retain high-quality demonstrations (``MQS + Filter'' in \autoref{tab:results-in-dataset} and \autoref{tab:retriever-performance}).
CodeS measures similarity scores of retrieved scripts based on both the masked questions and the original ones, taking the maximum of these two similarity scores (``MQS + QS'' in \autoref{tab:results-in-dataset} and \autoref{tab:retriever-performance}).
\DIFadd{Following the task definition in Section~\ref{ssec:task-def}, all retrieval operations are conducted within a database-specific historical query pool, i.e., for each question, the retriever only searches over historical SQL queries from the same database rather than across all databases.}

\DIFadd{To comprehensively evaluate \mydataset's difficulty, we additionally benchmark three other categories that do not inherently use dynamic retrieval\DIFdel{, evaluated in their original configurations}:
(1) Fine-tuned models~\cite{xiyan-sql, omnisql, csc-sql, slm-sql, sql-r1};
(2) Prompt-based frameworks~\cite{mac-sql, alphasql, ta-sql, din-sql, feather-sql} with zero-shot or static few-shot strategies;
(3) Foundation models like Qwen3 and GPT-5.3-Codex.}

\textbf{Apparatus}.
For the retrieval stage of our retriever, we employ the all-mpnet-base-v2\footnote{\url{https://huggingface.co/sentence-transformers/all-mpnet-base-v2}} model for embedding computation.
For the refinement stage, knowledge base construction, as well as SQL generation\DIFdel{in DAIL-SQL, OpenSearch-SQL, and N-rep pipelines}, we utilize the Qwen3~\cite{qwen3} model via API\footnote{Snapshot version: \href{https://www.alibabacloud.com/help/en/model-studio/models\#1e0e75495dyvu}{qwen-plus-2025-07-28}} as the underlying LLM.
For \DIFdel{CodeS, we use its fine-tuned 7B model deployed}\DIFadd{methods involving fine-tuned models}, we deployed them on an NVIDIA A100-SXM4-80GB GPU.
We set the LLM temperature to 0 across all tasks to keep deterministic outputs.
All experiments are conducted on a server running Ubuntu 22.04.5 LTS with 512 GB RAM and dual 64-core AMD EPYC 7763 CPUs (@ 2.45GHz, 256 threads total).

\textbf{Metrics}.\DIFdel{We evaluate our approach from two perspectives.
First, following the BIRD benchmark, we compute Execution Accuracy (EX), which measures the proportion of predicted SQL queries whose execution results match those of the ground-truth queries.
We compare the accuracy of each few-shot pipeline when using its original retriever versus our proposed retriever.
Second, we assess retrieval quality through precision, recall, and F1 score.}
\DIFadd{We evaluate Execution Accuracy (EX) for each text-to-SQL pipeline following the BIRD benchmark, and additionally measure retrieval quality using precision, recall, and F1 score.}
Since some knowledge items in BIRD development set appear in multiple scripts and some scripts contain multiple knowledge items, we calculate these metrics based on the number of correctly covered knowledge items rather than the number of correctly retrieved scripts.
Knowledge items are represented using SQL skeletons (with database values masked) to handle duplicate retrieval and ensure accurate metric computation.\DIFdel{As the number of required knowledge items varies across tasks, we compute overall precision and recall by aggregating true positives, false positives, and false negatives across all tasks.}
\DIFadd{We report overall precision and recall since the number of required knowledge items varies across tasks.}
\DIFdel{All metrics are reported as percentages.}

\subsection{Results}
\label{ssec:results}

\DIFadd{\textbf{Performance and Retrieval Quality}.
\autoref{tab:results-in-dataset} presents execution accuracy when integrating our retriever versus original retrievers across four text-to-SQL systems.
Our approach delivers consistent improvements, with DAIL-SQL benefiting most (\DIFdel{51.33\%, }+10.55\% over MQS).
Even the fine-tuned CodeS (7B) model improves by 6.03\%\DIFdel{despite lower absolute accuracy (18.95\%)}, suggesting that better retrievers can enhance resource-constrained deployments where large models may be impractical.
\autoref{tab:retriever-performance} shows our retriever achieves the highest F1\DIFdel{(49.77\%)} and recall\DIFdel{(83.07\%)}, successfully covering most required knowledge items.
However, modest precision (35.53\%) indicates excessive irrelevant knowledge still constrains accuracy, highlighting potential for more sophisticated filtering.
Notably, DAIL retriever's lowest F1 (28.69\%) stems from error propagation:
When initial MQS-based predictions lack external knowledge, subsequent skeleton-based filtering excludes relevant historical queries,
as evidenced by its 2.01\% accuracy decline versus MQS alone (\autoref{tab:results-in-dataset}).}

\DIFdel{\textbf{Overall Performance}.
\mbox{\autoref{tab:results-in-dataset}} presents the execution accuracy of different text-to-SQL systems on \mbox{\mydataset} when integrated with our proposed retriever versus their original retrievers.
Our approach demonstrates consistent improvements across all four pipelines.
Notably, DAIL-SQL benefits the most from our retriever, achieving 51.33\% accuracy, substantially outperforming MQS by 10.55\% and DAIL by 12.56\%.
CodeS (7B), unlike the other pipelines compatible with various foundation models (e.g., Qwen3), is a fine-tuned small text-to-SQL model.
Although its absolute execution accuracy (18.95\%) remains substantially lower than the Qwen3-based pipelines, our retriever still delivers a notable 6.03\% improvement over its original ``MQS + QS'' retriever (12.92\%).
This suggests that equipping fine-tuned text-to-SQL models with better retrievers holds considerable potential for performance improvement, offering a direction for resource-constrained deployment scenarios where LLM-based frameworks may be impractical.}

\begin{table}[t]
\centering
\small
\begin{tabularx}{\columnwidth}{lll}
    \toprule
    \textbf{Pipeline} & \textbf{Retriever} & \textbf{EX} \\
    \midrule
    \multirow{3}{*}{DAIL-SQL} & MQS & 40.78 \\
    & DAIL & 38.77 \\
    & Ours & \textbf{51.33} ($\uparrow$ 10.55) \\
    \midrule
    \multirow{2}{*}{OpenSearch-SQL} & MQS & 59.15 \\
    & Ours & \textbf{60.95} ($\uparrow$ 1.80) \\
    \midrule
    \multirow{2}{*}{N-rep} & MQS + Filter & 54.49 \\
    & Ours & \textbf{57.79} ($\uparrow$ 3.30) \\
    \midrule
    \multirow{2}{*}{CodeS (7B)} & MQS + QS & 12.92 \\
    & Ours & \textbf{18.95} ($\uparrow$ 6.03) \\
    \bottomrule
\end{tabularx}
\caption{Execution accuracy on \mydataset\ across different approaches.}
\label{tab:results-in-dataset}
\end{table}

\DIFdel{\textbf{Retrieval Quality Analysis}.
\mbox{\autoref{tab:retriever-performance}} compares the retrieval quality of our approach against the original retrievers used in existing pipelines.
Our retriever achieves the highest F1 score of 49.77\%, substantially outperforming all baselines, while demonstrating superior recall (83.07\%) that successfully covers the vast majority of required knowledge items across tasks.
However, the relatively modest precision (35.53\%) suggests that excessive irrelevant knowledge may still constrain execution accuracy, indicating potential for future improvement through more sophisticated filtering mechanisms.
Notably, DAIL retriever exhibits the lowest performance (28.69\% F1), likely due to error propagation in its two-stage mechanism:
when the initial SQL prediction based on MQS lacks external knowledge,
subsequent skeleton-based filtering may exclude relevant historical queries,
as evidenced by its 2.01\% execution accuracy decline compared to MQS only in \mbox{\autoref{tab:results-in-dataset}}.}

\begin{table}[t]
\centering
\small
\begin{tabularx}{\columnwidth}{Xccc}
    \toprule
    \textbf{Retriever} & \textbf{Precision} & \textbf{Recall} & \textbf{F1} \\
    \midrule
    MQS & 21.11 & 48.84 & 29.48 \\
    DAIL & 20.44 & 48.10 & 28.69 \\
    MQS + Filter & \underline{25.09} & \underline{57.18} & \underline{34.88} \\
    MQS + QS & 24.38 & 52.47 & 33.29 \\
    \midrule
    Ours & \textbf{35.53} & \textbf{83.07} & \textbf{49.77} \\
    \bottomrule
\end{tabularx}
\caption{Retrieval results on \mydataset\ across different retriever in existing Text-to-SQL pipelines.}
\label{tab:retriever-performance}
\end{table}

\DIFadd{\textbf{Extended Baseline Comparison}.
As shown in \autoref{tab:other_pipelines}, these systems achieve 11\%-46\% execution accuracy on \mydataset\ in their original configurations.
According to \autoref{tab:foundation_model}, the five foundation models achieve 16\%-34\% accuracy despite their strong performance on BIRD development set (35\%-52\%).
These results confirm that our benchmark poses substantial challenges that existing pipelines and LLMs cannot easily overcome without effective history-grounded mechanisms.
Detailed results are provided in Appendix~\ref{appssec:extended_baseline_comparison}.}

\DIFadd{\textbf{Ablation on Retrieval Granularity and Knowledge}.
To better understand the impact of retrieval granularity and the necessity of historical knowledge, we conduct additional ablations on the OpenSearch-SQL, the best-performing pipeline at 60.95\% in \autoref{tab:results-in-dataset}.
As shown in \autoref{tab:ablation_retrieval_levels} of Appendix~\ref{appssec:retr_gran_know}, both query-level and component-level retrieval individually improve execution accuracy over the no-retrieval setting, while their combination achieves the best performance.
Disabling retrieval leads to a substantial performance drop, while replacing retrieved results with gold annotations yields significant gains, jointly confirming that leveraging accurate historical knowledge is essential for \mydataset.}

\DIFadd{\textbf{Retrieval Performance by Knowledge Type}.
Table~\ref{tab:retrieval_by_knowtype} breaks down retrieval quality across the five knowledge types defined in Section~\ref{sec:baseline-approach}. 
\textit{Condition} knowledge achieves the highest precision\DIFdel{(36.70\%)} and F1\DIFdel{(49.64\%)}, as filtering predicates (e.g., \texttt{publisher\_name = 'Marvel Comics'}) are typically explicit in questions. 
\textit{Output} knowledge attains the highest recall (84.26\%) since components like \texttt{LIMIT} and \texttt{DISTINCT} are rarely missed, though their ubiquity inflates false positives.
In contrast, \textit{relation} and \textit{calculation} knowledge show lower precision (both below 30\%), likely because natural language questions rarely specify exact JOIN paths or aggregation formulas explicitly.}
\DIFdel{These patterns suggest future work should prioritize improving precision for implicit knowledge types through schema-aware matching or query intent modeling.}

\begin{table}[ht]
\centering
\small
\begin{tabularx}{\columnwidth}{Xccc}
\toprule
\textbf{Type} & \textbf{Precision} & \textbf{Recall} & \textbf{F1} \\
\midrule
Condition & \textbf{36.70} & 76.67 & \textbf{49.64} \\
Output & 33.62 & \textbf{84.26} & 48.06 \\
Dimension & 30.24 & 79.19 & 43.77 \\
Relation & 25.75 & 78.09 & 38.73 \\
Calculation & 23.33 & 76.44 & 35.75 \\
\bottomrule
\end{tabularx}
\caption{Retrieval results for each knowledge type using OpenSearch-SQL pipeline with our retriever.}
\label{tab:retrieval_by_knowtype}
\end{table}

\DIFadd{\textbf{Impact of Shot Selection Strategy}.
\DIFdel{Using OpenSearch-SQL, the best-performing pipeline at 60.95\% in \mbox{\autoref{tab:results-in-dataset}}, we}\DIFadd{We} analyze how the maximum shot parameter affects retrieval quality and generation accuracy \DIFadd{using OpenSearch-SQL}.
\autoref{tab:topk-performance} shows execution accuracy peaks at 61.31\% with 4 shots (average 3.744 actual shots), slightly outperforming the dynamic approach (60.95\%, average 4.572 shots), suggesting overly permissive retrieval introduces noise.
As max shots increase from 1 to dynamic, precision drops from 57.95\% to 35.53\% while recall rises from 42.78\% to 83.07\%.
Despite declining precision, the dynamic strategy maintains competitive accuracy, indicating the LLM-based refinement stage effectively filters irrelevant knowledge.}

\DIFdel{\textbf{Impact of Shot Selection Strategy}.
Given that OpenSearch-SQL achieves the highest execution accuracy (60.95\%) in \mbox{\autoref{tab:results-in-dataset}}, we use this pipeline to analyze the sensitivity of the maximum shot parameter on both retrieval quality and generation accuracy.
\mbox{\autoref{tab:topk-performance}} shows that execution accuracy peaks at 61.31\% when limiting retrieval to 4 shots (average 3.744 actual shots), slightly outperforming the dynamic approach (60.95\%, average 4.572 shots).
This suggests that overly permissive retrieval may introduce noise that degrades generation quality.
The trend in F1 scores reveals an expected trade-off:
as max shots increase from 1 to dynamic, precision drops from 57.95\% to 35.53\% while recall rises from 42.78\% to 83.07\%.
Despite the decline in precision, the dynamic strategy maintains competitive execution accuracy, indicating that the LLM-based refinement stage effectively filters less relevant knowledge.}

\begin{table}[ht]
\centering
\resizebox{\columnwidth}{!}{
\begin{tabular}{cccccc}
    \toprule
    \textbf{Max shot} & \textbf{Avg shot} & \textbf{EX} & \textbf{Precision} & \textbf{Recall} & \textbf{F1} \\
    \midrule
    1 & 1.000 & 58.36 & \textbf{57.95} & 42.78 & 49.23 \\
    2 & 1.999 & 59.30 & 48.37 & 62.13 & \textbf{54.39} \\
    3 & 2.967 & 59.58 & 42.55 & 73.14 & 53.80 \\
    4 & 3.744 & \textbf{61.31} & 38.75 & 79.33 & 52.07 \\
    5 & 4.229 & 60.01 & 36.68 & 81.60 & 50.61 \\
    Dynamic & 4.572 & 60.95 & 35.53 & \textbf{83.07} & 49.77 \\
    \bottomrule
\end{tabular}
}
\caption{Execution accuracy and retrieval results of our proposed retriever with different max shots on the OpenSearch-SQL pipeline.}
\label{tab:topk-performance}
\end{table}

\DIFadd{\textbf{Effect of Similarity Threshold}.
Similar to \autoref{tab:topk-performance}, we analyze how similarity thresholds in the initial retrieval stage affect both retrieval quality and generation performance using OpenSearch-SQL.
\autoref{tab:threshold-performance} shows\DIFdel{optimal execution accuracy (60.73\%) at threshold 0.6, closely followed by 0.7 (60.66\%). Lower}
\DIFadd{that lower} thresholds (0.5) maintain high recall (82.52\%) but introduce noise, while higher thresholds (0.8-0.9) prematurely drop recall\DIFdel{ to 54.94\% and 32.68\%}.
The threshold of 0.6 achieves favorable balance\DIFdel{ (81.33\% recall, 35.78\% precision)} consistent with our two-stage design:
the embedding-based stage casts a wide net for candidates, while LLM-based reranking performs fine-grained filtering.
Aggressive thresholds exclude queries the LLM could identify through deeper understanding;
permissive thresholds introduce noise interfering with LLM judgment\DIFdel{, evidenced by the sharp drop at 0.9 (56.57\% EX, 33.07\% F1)}.}

\DIFdel{\textbf{Effect of Similarity Threshold}.
Similar to \mbox{\autoref{tab:topk-performance}}, we analyze the impact of similarity threshold in our retriever's initial retrieval stage on both retrieval quality and downstream generation performance using the OpenSearch-SQL pipeline.
\mbox{\autoref{tab:threshold-performance}} shows that the optimal execution accuracy of 60.73\% is achieved at a threshold of 0.6, closely followed by 60.66\% at 0.7.
The recall metric reveals the underlying mechanism:
at lower thresholds (i.e., 0.5), recall remains high (82.52\%) but introduces excessive noise, while at a higher thresholds (i.e., 0.8-0.9), recall drops dramatically to 54.94\% and 32.68\% respectively, filtering out too much relevant knowledge.
The threshold of 0.6 achieves a favorable balance (81.33\% recall, 35.78\% precision) consistent with our two-stage retrieval design: 
the initial embedding-based stage casts a wide net to provide sufficient candidate space, while the subsequent LLM-based reranking performs fine-grained semantic filtering.
Excessively aggressive thresholds prematurely exclude queries that the LLM could identify through deeper semantic understanding, whereas excessively permissive thresholds introduce noise that interferes with the LLM's judgment, as evidenced by the sharp performance drop at 0.9 (56.57\% EX, 33.07\% F1).}

\begin{table}[ht]
\centering
\small
\begin{tabularx}{\columnwidth}{Xcccc}
    \toprule
    \textbf{Similarity} & \textbf{EX} & \textbf{Precision} & \textbf{Recall} & \textbf{F1} \\
    \midrule
    0.5 & 58.79 & 35.58 & \textbf{82.52} & \textbf{49.72} \\
    0.6 & \textbf{60.73} & \textbf{35.78} & 81.33 & 49.70 \\
    0.7 & 60.66 & 35.17 & 74.42 & 47.77 \\
    0.8 & 59.15 & 34.19 & 54.94 & 42.15 \\
    0.9 & 56.57 & 33.48 & 32.68 & 33.07 \\
    \bottomrule
\end{tabularx}
\caption{Execution accuracy and retrieval results of our retriever with different similarity thresholds on the OpenSearch-SQL pipeline.}
\label{tab:threshold-performance}
\end{table}

\DIFadd{\textbf{Latency and Storage}.
See Appendix~\ref{ssec:efficiency-analysis} for detailed analysis of our proposed retriever.}

\subsection{Case Studies}

\DIFadd{To understand how retrieval quality affects generation, we examine two representative cases with 100\% recall but different outcomes.}

\DIFadd{\textbf{Success Case} (question\_id=1145, ``student\_club'' database):
The task requires joining \texttt{member} and \texttt{zip\_code} tables to query hometown information.
The retriever successfully identified a historical query containing the non-obvious foreign key relationship 
\texttt{member.zip} $\rightarrow$ \texttt{zip\_code.zip\_code}, 
which cannot be inferred from column names alone.
With 90\% precision and 100\% recall, the LLM correctly generated the required join pattern.}

\DIFadd{\textbf{Failure Case} (question\_id=1213, ``superhero'' database):
Despite 100\% recall, the retriever returned 14 candidates including both the correct subquery pattern (\texttt{WHERE attribute\_value = (SELECT MAX(...))}) and an oversimplified alternative (\texttt{ORDER BY ... DESC LIMIT 1}).
With only 28.6\% precision, the LLM selected the latter, which returns one result instead of all heroes tied at maximum strength.
This demonstrates that high recall with low precision can mislead generation when multiple plausible patterns coexist.
Detailed SQL comparisons are provided in Appendix~\ref{appssec:case_studies}.}

\subsection{Discussion}

\DIFadd{We identify three potential causes for the modest precision of our retriever (35.53\% in \autoref{tab:retriever-performance})}.

\DIFadd{\textbf{Over-retrieval of Generic Fragments}.
Our baseline method (Section~\ref{sec:baseline-approach}) maps natural language keywords (including generic ones like ``top 1'' and ``unique'') to fine-grained SQL components like \texttt{LIMIT 1} (10,619 occurrences) and \texttt{DISTINCT} (6,163 occurrences) to maximize recall.
However, tracing these generic components back to their source queries inevitably retrieves semantically irrelevant candidates and inflates false positives.}

\DIFadd{\textbf{Insufficient Semantic Discernment}.
Our approach lacks the granularity to distinguish subtle semantic differences between highly similar fragments, such as \texttt{CAST(satscores.NumGE1500 AS REAL) / satscores.NumTstTakr} versus \texttt{ORDER BY CAST(T1.NumGE1500 AS REAL) / T1.NumTstTakr ASC}.
Consequently, to safeguard recall, our approach retains all such candidates rather than effectively filtering or deduplicating them, which contributes to high false positive rates.}

\DIFadd{\textbf{Neglect of Co-occurrence Patterns}.
Our retrieval method treats each SQL fragment independently without modeling deeper associative knowledge, such as the frequent co-occurrence between specific conditions and relations.
Leveraging such co-occurrence patterns as soft constraints could effectively filter out incompatible fragments, yet their absence allows many semantically irrelevant candidates to pass through.}

\section{Conclusion}
\DIFdel{In this paper, we introduce \mbox{\mydataset}, a benchmark comprising 1,393 tasks across 11 databases built upon BIRD to evaluate text-to-SQL systems' ability to leverage historical queries for SQL generation.
We propose a plug-in retriever as a baseline approach that extracts five types of external knowledge and achieves consistent improvements across four representative text-to-SQL pipelines.}
\DIFadd{In this paper, we introduce \mydataset\ to benchmark history-aware text-to-SQL generation and validate its utility via a plug-in retriever that yields consistent improvements across standard pipelines.}
Future work could explore more sophisticated filtering mechanisms to reduce irrelevant knowledge in the retrieval results.
Alternative knowledge categorization schemes that better align with LLM reasoning capabilities may further improve generation quality.
Moreover, future research could also investigate representing retrieved knowledge through structured descriptions or relevant SQL fragments rather than complete historical queries to potentially enhance the effectiveness of LLM prompts.
We hope \mydataset\ advances research on bridging the gap between benchmark evaluation and real-world data querying scenarios.

\section*{Limitations}
We identify two limitations of our work.
First, our benchmark \DIFdel{treats historical query logs as static by using the fixed BIRD development set, whereas real-world query logs evolve continuously as users submit new queries.}\DIFadd{models a controlled scenario where we retain only final, successful SQL queries from past sessions, using the fixed BIRD development set as this collection.
This differs from raw, session-based query logs that include failed attempts, syntax errors, and iterative refinements.
While real-world logs may contain such noise and evolve continuously as users submit new queries, our design choice allows for systematic evaluation of knowledge retrieval and utilization without the confounding effects of malformed queries.
However, this} assumption may not capture how the expanding retrieval pool and shifting knowledge patterns over time might influence retrieval performance and generation quality.
\DIFadd{Future work could incorporate more realistic, noisy query logs to bridge this gap.}
Second, the modest precision of our retriever indicates that the current retrieval approach may introduce irrelevant knowledge alongside useful historical patterns, which contributes to the moderate execution accuracy observed in our experiments.

\section*{Ethical Consideration}
Our dataset construction process involved three authors with SQL expertise who conducted validation of all generated tasks to ensure data quality.
No crowdsourcing was involved, and no personal or sensitive information was collected.
We leverage LLMs to help generate natural language questions and descriptions during the dataset construction process. 
We acknowledge that LLMs may generate biased or inappropriate content.
To mitigate this risk, the three authors carefully reviewed all LLM-generated content to ensure semantic consistency, task naturalness, and the absence of offensive language in our dataset.

\section*{Acknowledgement}
This work was supported by Ningbo Yongjiang Talent Programme (2024A-399-G).
We sincerely appreciate the constructive feedback from the anonymous reviewers.

 \bibliography{custom}

\appendix

 \clearpage 

\section{Dataset}
\subsection{Dataset Description}

Similar to BIRD~\cite{bird}, \mydataset\ is in JSON format and each task contains the fields ``question\_id'', ``db\_id'', ``question'', ``SQL'', and ``history\_knowledge''.
\autoref{fig:task-example} shows an example task in \mydataset.
The first four fields follow the convention in the BIRD dataset:

\begin{itemize}[itemsep=0pt, parsep=0pt]
    \item \textbf{question\_id}: A unique integer identifier for each task.
    \item \textbf{db\_id}: The database ID.
    \item \textbf{question}: The natural language query.
    \item \textbf{SQL}: The ground-truth SQL query that correctly answers the question.
\end{itemize}

We introduce an additional field ``\textbf{history\_knowledge}'' to enable systematic evaluation of retrieval effectiveness.
This field maps SQL knowledge fragments (represented as skeletons with database-specific values masked using \texttt{<val>} placeholders) to the question IDs from the BIRD development set where these fragments appear.
Each key in ``history\_knowledge'' represents a specific piece of external knowledge required to generate the target SQL query, while the corresponding value is a list of question IDs indicating which historical queries contain this knowledge.

For instance, in \autoref{fig:task-example}, the entry \texttt{"schools.Charter=<val>":[60,61,62,63,65]}
indicates that the filtering condition \texttt{"schools.Charter=<val>"} (which could be instantiated as \texttt{schools.Charter=1} or \texttt{schools.Charter=0} depending on context) can be found in historical queries with question IDs 60, 61, 62, 63, or 65 from the BIRD development set.
Similarly, \texttt{schools.FundingType='<val>':[28]} shows that knowledge about filtering by funding type appears in question 28.
This fine-grained annotation enables precise evaluation of whether retrieval systems successfully identify the specific historical queries and SQL fragments needed to solve each task.

\begin{figure*}[ht]
  \includegraphics[width=\textwidth]{figtex/figures/TaskExample_v1_8x.jpg}
  \caption{A task example in \mydataset.}
  \label{fig:task-example}
\end{figure*}

Following the evaluation methodology described in Section~\ref{ssec:settings}, we calculate retrieval metrics based on the number of knowledge items covered rather than the number of SQL scripts retrieved.
This design choice reflects the fact that a single historical SQL script may contain multiple relevant knowledge fragments, and multiple scripts may contain the same knowledge item.
For instance, in the example shown in \autoref{fig:task-example}, suppose a retrieval system returns historical queries with question IDs 65 and 87.
According to the ground-truth annotations in ``history\_knowledge'', query 65 contains the knowledge fragment \texttt{schools.Charter = <val>},
while query 87 contains both \texttt{FROM frpm AS T1 INNER JOIN schools AS T2 ON T1.CDSCode = T2.CDSCode} and \texttt{schools.County = '<val>'}.
Although only 2 scripts are retrieved, this corresponds to 3 distinct knowledge items being successfully covered out of the 4 required knowledge items annotated for this task (the join pattern, county filtering, and charter filtering).
The recall of this task would therefore be calculated as 3/4 = 75\%.
For more details please refer to our repository:
\url{https://github.com/zjuidg/BIRD-History}.

\subsection{Knowledge Type Categorization}
\label{appssec:knowledge_type_cate}

\begin{table*}[t]
\centering
\small
\begin{tabularx}{\textwidth}{lllX}
\toprule
\textbf{Type} & \textbf{SQL Clause} & \textbf{Example SQL Components} & \textbf{Description} \\
\midrule
\multirow{2}{*}{Calculation} & \multirow{2}{*}{\makecell[l]{\texttt{SELECT},\\\texttt{ORDER BY}}} & \makecell[l]{\lstinline{CAST(COUNT(CASE WHEN T.element = 'cl'} \\ \lstinline{THEN T.atom_id ELSE NULL END) AS REAL)} \\ \lstinline{ * 100 / COUNT(T.atom_id)}} & \makecell[l]{Chlorine atom percentage among all atoms\\ in molecules} \\
\cmidrule{3-4}
& & \makecell[l]{
\lstinline{ORDER BY SUM(T1.home_team_goal} \\ \lstinline{+ T1.away_team_goal)}} & \makecell[l]{Total goals across all matches in a \\ football team} \\
\midrule
\multirow{2}{*}{Condition} & \multirow{2}{*}{\makecell[l]{\texttt{WHERE},\\\texttt{HAVING}}} & 
\lstinline{SUBSTRING(T1.event_date,1,4)='2019'} & Filter the events in the year 2019 \\
\cmidrule{3-4}
& & \lstinline{molecule.label = '-'} & Filter non-carcinogenic molecules \\
\midrule
Relation & \texttt{FROM}, \texttt{JOIN} & \makecell[l]{\lstinline{FROM atom T1 INNER JOIN molecule T2} \\ \lstinline{ON T1.molecule_id = T2.molecule_id}} & \makecell[l]{Should join ``atom'' and ``molecule'' \\ tables to obtain labels for atoms} \\
\midrule
Dimension & \texttt{GROUP BY} & \lstinline{GROUP BY sname, charter_funding_type} & \makecell[l]{Should consider both student and funding \\ when aggregating average math scores} \\
\midrule
\multirow{2}{*}{Output} & \multirow{2}{*}{\makecell[l]{\texttt{ORDER BY},\\\texttt{LIMIT},\\\texttt{DISTINCT}}} & 
\lstinline{ORDER BY result DESC LIMIT 10} & Return top 10 groups sorted by score \\
\cmidrule{3-4}
& & \lstinline{SELECT DISTINCT category, type} & \makecell[l]{Find the distinct budget categories \\ and event types} \\
\bottomrule
\end{tabularx}
\caption{Categorization of external knowledge in SQL queries by grammar. These five types cover fundamental SQL clauses, and complex structures such as subqueries or common table expressions (CTEs) can be recursively parsed to extract the same knowledge types.}
\label{tab:knowledge_type}
\end{table*}

\DIFadd{According to Section~\ref{sec:baseline-approach}, we categorize external knowledge embedded in historical SQL queries into five types based on their syntactic positions in SQL grammar, following SQLite specifications \footnote{\url{https://www.sqlite.org/index.html}}.
\autoref{tab:knowledge_type} presents the complete categorization with representative examples from BIRD benchmark:}

\begin{itemize}[itemsep=0pt, parsep=0pt]
\item \DIFadd{\textit{Calculation} knowledge captures computational logic through formulas and functions in \texttt{SELECT} and \texttt{ORDER BY} clauses, ranging from simple arithmetic operations to complex aggregations like percentage calculations.}
\item \DIFadd{\textit{Condition} knowledge encodes filtering predicates in \texttt{WHERE} and \texttt{HAVING} clauses, including both simple equality checks and sophisticated pattern matching operations (e.g., \texttt{SUBSTRING} for temporal filtering).}
\item \DIFadd{\textit{Relation} knowledge specifies table join patterns in \texttt{FROM}-\texttt{JOIN} clauses, documenting how entities are connected through foreign key relationships.
This is valuable as join paths are often non-obvious from schema alone.}
\item \DIFadd{\textit{Dimension} knowledge determines aggregation granularity via \texttt{GROUP BY} clauses, indicating which attributes should be considered when computing group-level statistics.}
\item \DIFadd{\textit{Output} knowledge defines result presentation preferences through \texttt{ORDER BY}, \texttt{LIMIT}, and \texttt{DISTINCT} clauses, such as returning top-k results or ensuring uniqueness of records.}
\end{itemize}

\DIFadd{These five types comprehensively cover fundamental SQL constructs. For complex structures such as subqueries, common table expressions (CTEs), or window functions, our parser recursively applies the same categorization to nested components, ensuring complete knowledge coverage regardless of query complexity.}

\subsection{Dataset Analysis}
\label{appssec:dataset-analysis}

\DIFadd{To systematically characterize \mydataset's task complexity and knowledge diversity, we analyze four complementary dimensions:
Benchmark comparison, knowledge item count, knowledge type diversity, and knowledge type distribution.}

\DIFadd{\textbf{Benchmark comparison}.
\autoref{tab:comparison} positions \mydataset\ among existing text-to-SQL benchmarks.
While Spider 2.0 datasets exhibit higher token counts due to enterprise-scale schemas, \mydataset\ achieves the highest average JOIN length (2.57 tables on average) and nesting depth (2.06 layers on average), indicating superior reasoning complexity.
Critically, \mydataset\ is the only benchmark providing knowledge retrieval annotations, enabling evaluation of how models identify and compose reusable patterns from historical queries.
Our complexity arises naturally from knowledge composition (averaging 5.3 knowledge items per task) rather than artificially inflated SQL syntax, reflecting realistic analyst workflows.}

\begin{table*}[ht]
\centering
\small
\begin{tabularx}{\textwidth}{l*{11}{>{\centering\arraybackslash}X}}
\toprule
\textbf{Knowledge Items} & \textbf{1} & \textbf{2} & \textbf{3} & \textbf{4} & \textbf{5} & \textbf{6} & \textbf{7} & \textbf{8} & \textbf{9} & \textbf{10} & \textbf{11} \\
\midrule
\textbf{\# Tasks} & 20 & 95 & 156 & 213 & 262 & 253 & 192 & 123 & 53 & 17 & 9 \\
\bottomrule
\end{tabularx}
\caption{\DIFadd{Distribution of knowledge item count per task.}}
\label{tab:distribution_knowledge_count}
\end{table*}

\DIFadd{\textbf{Knowledge item count per task}.
\autoref{tab:distribution_knowledge_count} presents the distribution of knowledge items required per task.
The distribution peaks at 5 items (262 tasks, 18.8\%),
with 1,278 tasks (91.7\%) requiring 3 or more knowledge items.
This demonstrates that the majority of \mydataset\ tasks demand reasoning over multiple historical queries rather than simple one-to-one knowledge reuse.
The long tail (tasks requiring 8-11 items) represents particularly challenging cases where the target query synthesizes domain knowledge from numerous historical queries, reflecting realistic scenarios where analysts build complex queries by incrementally combining proven patterns.}

\begin{table}[ht]
\centering
\small
\begin{tabularx}{\columnwidth}{l*{6}{>{\centering\arraybackslash}X}}
\toprule
\textbf{Knowledge Types} & \textbf{1} & \textbf{2} & \textbf{3} & \textbf{4} & \textbf{5} & \textbf{6} \\
\midrule
\textbf{\# Tasks} & 100 & 320 & 568 & 274 & 127 & 4 \\
\bottomrule
\end{tabularx}
\caption{\DIFadd{Distribution of distinct knowledge type count per task.}}
\label{tab:distribution_knowledge_type}
\end{table}

\DIFadd{\textbf{Knowledge type diversity per task}. 
\autoref{tab:distribution_knowledge_type} analyzes how many distinct knowledge types each task requires,
where types include relation (table joins), condition (filtering predicates), dimension (aggregation granularity), calculation (formulas and functions), output (result specification), and full SQL (complete query reuse).
The distribution peaks at 3 types (568 tasks, 40.8\%), with 973 tasks (69.8\%) requiring 3 or more distinct types.
This cross-type integration complexity distinguishes \mydataset\ from benchmarks like Spider-DK~\cite{spider-dk} where tasks predominantly focus on single knowledge categories.
The prevalence of multi-type tasks validates that real-world SQL authoring requires coordinated reasoning across multiple semantic dimensions.}

\begin{table}[ht]
\centering
\small
\begin{tabularx}{\columnwidth}{l*{3}{>{\centering\arraybackslash}X}}
\toprule
\textbf{Type} & \textbf{Total Items} & \textbf{Proportion (\%)} & \textbf{Avg Items per Task} \\
\midrule
Relation & 1,318 & 16.6 & 0.946 \\
Condition & 4,002 & 50.3 & 2.873 \\
Dimension & 244 & 3.1 & 0.175 \\
Calculation & 1,004 & 12.6 & 0.721 \\
Output & 1,354 & 17.0 & 0.972 \\
Full SQL & 29 & 0.4 & 0.021 \\
\bottomrule
\end{tabularx}
\caption{\DIFadd{Total knowledge items, proportions, and average knowledge items per task for each knowledge type.}}
\label{tab:stat_know_type}
\end{table}

\DIFadd{\textbf{Knowledge type distribution}. 
\autoref{tab:stat_know_type} breaks down the prevalence of each knowledge type across all tasks.
Condition knowledge dominates (4,002 items, 50.3\% of all knowledge items), reflecting the central role of filtering predicates in SQL queries.
Output specifications (1,354 items, 17.0\%) and relation patterns (1,318 items, 16.6\%) are also prevalent, indicating that result formatting and join logic constitute challenges in history-driven text-to-SQL.
The rare full SQL category (29 items, 0.4\%) demonstrates that the tasks of \mydataset\ rarely permit direct reuse of entire historical queries.
Instead, most tasks require extracting and composing specific patterns from multiple sources, significantly increasing the difficulty beyond simple query template matching.}

\subsection{License and Usage Compliance}

\DIFadd{This section addresses the licensing requirements and usage compliance for the dataset (i.e., BIRD) used in our research.}

\DIFadd{\textbf{Data License Information}.
The BIRD dataset is released under the CC BY-SA 4.0 license. This license requires that:
(1) Proper attribution must be given to the original authors;
(2) Any derivative works must be distributed under the same license terms (share-alike requirement).}

\DIFadd{\textbf{Usage Consistency}.
Our use of the BIRD dataset is fully consistent with its intended research purposes:
(1) We maintain the original CC BY-SA 4.0 license for our derivative dataset, \mydataset;
(2) The dataset is used exclusively for academic research purposes;
(3) We provide appropriate attribution to the original BIRD dataset creators;
(4) Our derivative work complies with the share-alike requirement by also being released under CC BY-SA 4.0.}

\section{Additional Analysis of Experiment Results}
\subsection{Extended Baseline Comparison}
\label{appssec:extended_baseline_comparison}

\DIFadd{To further highlight the difficulty of \mydataset, we compare performance of the same systems on the original BIRD development set (BIRD Dev) and \mydataset.
We assess three additional categories of text-to-SQL approaches beyond the four few-shot systems discussed in Section~\ref{ssec:results}.}

\begin{table}[ht]
\centering
\small
\begin{tabularx}{\columnwidth}{l*{3}{>{\centering\arraybackslash}X}}
\toprule
\textbf{Method} & \textbf{\makecell[c]{EX on\\BIRD-History}} & \textbf{\makecell[c]{EX on\\BIRD Dev}} \\
\midrule
\multicolumn{3}{c}{\textbf{Finetuned Models}} \\
XiYanSQL-32B & 22.97 & 67.01 \\
OmniSQL-32B & 19.02 & 69.23 \\
CscSQL-7B & 26.49 & 69.19 \\
SLM-SQL-1.5B & 21.75 & 67.08 \\
SQL-R1-14B & 28.64 & 67.14 \\
\midrule
\multicolumn{3}{c}{\textbf{Prompt-based Frameworks}} \\
MAC-SQL & 46.37 & 47.91 \\
Alpha-SQL & 44.72 & 56.52 \\
TA-SQL & 31.08 & 50.98 \\
DIN-SQL & 10.91 & 27.77 \\
Feather-SQL & 27.57 & 39.90 \\
\bottomrule
\end{tabularx}
\caption{Performance of two additional categories of text-to-SQL pipelines on \mydataset\ and BIRD development set. Here, ``XiYanSQL-32B'' refers to \texttt{XiYanSQL-QwenCoder-32B-2504}, ``CscSQL-7B'' refers to \texttt{CscSQL-Qwen2.5-Coder-7B-Instruct}.}
\label{tab:other_pipelines}
\end{table}

\DIFadd{\textbf{Finetuned Models}.
\autoref{tab:other_pipelines} presents the execution accuracy of five finetuned models on both \mydataset\ and BIRD development set.
While these models achieve competitive performance on BIRD Dev (ranging from 67.01\% to 69.23\%), their accuracy drops drastically on \mydataset, ranging from merely 19.02\% (OmniSQL-32B) to 28.64\% (SQL-R1-14B).
The relatively modest performance suggests that without an effective retrieval mechanism to access historical context,
even models finetuned on text-to-SQL datasets struggle to resolve the underspecified queries characteristic of \mydataset.}

\DIFadd{\textbf{Prompt-based Frameworks}. \autoref{tab:other_pipelines} shows the performance of five prompt-based text-to-SQL frameworks that employ different prompting strategies without requiring model finetuning.
To ensure a fair comparison, the underlying foundation model for all these frameworks is consistent with the apparatus mentioned in Section~\ref{ssec:settings}, utilizing the Qwen3 API (qwen-plus-2025-07-28).
Similar to the finetuned models, these frameworks exhibit a consistent decline in execution accuracy when shifting from BIRD Dev to \mydataset.
MAC-SQL achieves the highest execution accuracy (46.37\%), followed by Alpha-SQL (44.72\%). 
Notably, DIN-SQL achieves only 10.91\% accuracy, likely due to its reliance on static few-shot examples that may not align well with the domain-specific knowledge patterns in \mydataset.}

\begin{table}[ht]
\centering
\small
\setlength{\tabcolsep}{1pt}
\begin{tabularx}{\columnwidth}{l*{2}{>{\centering\arraybackslash}X}}
\toprule
\textbf{Model} & \textbf{\mydataset} & \textbf{BIRD Dev} \\
\midrule
Qwen3 & 26.42 & 44.00 \\
DeepSeek-V3.2 & 21.11 & 35.72 \\
Llama3.3-70B-Instruct & 16.65 & 35.07 \\
GPT-5.3-Codex & 28.93 & 41.26 \\
Gemini-3-Flash & 33.24 & 51.56 \\
\bottomrule
\end{tabularx}
\caption{Execution accuracy comparison of different foundation models on \mydataset\ and BIRD development benchmarks.}
\label{tab:foundation_model}
\end{table}

\DIFadd{\textbf{Foundation models}.
\autoref{tab:foundation_model} presents execution accuracy comparisons, revealing substantial performance variations across models.
On BIRD-History, Gemini-3-Flash achieves the best performance (33.24\%), while Llama3.3-70B-Instruct demonstrates modest results (16.65\%) despite having fewer parameters.
The performance gap between \mydataset\ and BIRD Dev highlights the additional complexity, with Llama3.3-70B-Instruct showing the largest discrepancy (16.65\% vs. 35.07\%).
These results echo the findings from finetuned models:
even powerful foundation models face substantial difficulties without effective retrieval mechanism.}

\subsection{Retrieval Granularity and Knowledge}
\label{appssec:retr_gran_know}

\DIFadd{We provide additional ablation studies to further analyze the role of retrieval granularity and knowledge annotations.}

\DIFadd{\textbf{Retrieval Granularity}.
We compare three retrieval strategies:
(1) query-level retrieval (Q),
(2) component-level retrieval (C), and
(3) their combination (Q+C).
The results in \autoref{tab:ablation_retrieval_levels} show that while both Q and C improve over the no-retrieval setting, combining them achieves the best performance. 
This indicates that component-level retrieval captures fine-grained knowledge that is not fully covered by query-level similarity.}

\DIFadd{\textbf{Necessity of Retrieval.}
We include a ``None'' setting where the retriever is disabled (\autoref{tab:ablation_retrieval_levels}). 
The significant performance drop demonstrates that historical knowledge is essential for solving BIRD-History tasks.}

\DIFadd{\textbf{Effectiveness of Annotations.}
We further evaluate a ``Gold'' setting, where retrieved knowledge is replaced with ground-truth annotations (\autoref{tab:ablation_retrieval_levels}).
The performance gain suggests that the annotated \texttt{history\_knowledge} provides useful supervision for generation, and that improving retrieval accuracy remains a key direction.}

\begin{table}[ht]
\centering
\small
\begin{tabularx}{\columnwidth}{l*{4}{>{\centering\arraybackslash}X}}
\toprule
\textbf{Level} & \textbf{EX} & \textbf{Precision} & \textbf{Recall} & \textbf{F1} \\
\midrule
Q+C   & 60.95 & 35.53 & 83.07 & 49.77 \\
Q     & 59.44 & 34.61 & 69.45 & 46.19 \\
C     & 60.73 & 33.91 & 79.29 & 47.50 \\
None  & 54.27 & 0     & 0     & 0     \\
Gold  & 67.34 & 100   & 100   & 100   \\
\bottomrule
\end{tabularx}
\caption{Ablation studies across different retrieval levels on the best-performing few-shot pipeline, OpenSearch-SQL. ``Q'' denotes query-level retrieval, ``C'' denotes component-level retrieval, and ``Q+C'' refers to our proposed retriever that combines both levels. ``None'' indicates no retriever, while ``Gold'' represents oracle retrieval results.}
\label{tab:ablation_retrieval_levels}
\end{table}

\subsection{Efficiency Analysis}
\label{ssec:efficiency-analysis}

We conducted efficiency analysis of our proposed retriever (\autoref{tab:efficiency_analysis}) under the same environment described in Section~\ref{ssec:settings} Apparatus:
Embeddings are computed with \texttt{all-mpnet-base-v2}, and the LLM-based stages use Qwen3 through the API \texttt{qwen-plus-2025-07-28} with temperature 0 on the same Ubuntu 22.04.5 server with 512 GB RAM and dual 64-core AMD EPYC 7763 CPUs.

\textbf{Offline Indexing}.
The indexing stage consists of SQL parsing, natural language description generation, and embedding computation.
Among these, description generation dominates the cost, as it involves LLM calls.
However, this process is performed only once during knowledge base construction and does not affect inference latency.

\textbf{Online Retrieval.}
For each query, the retriever performs keyword extraction, embedding-based retrieval, and LLM-based filtering and reranking.
The embedding-based retrieval is efficient (about 54 ms), while LLM-based steps account for most of the latency.

\textbf{Storage Cost.}
Each SQL query requires approximately 20.62 KB to store its structured components, descriptions, and embeddings, resulting in moderate storage overhead.

\textbf{Discussion.}
The reported latency is measured in our current experimental setup using serial \texttt{qwen-plus-2025-07-28} API calls, as described in Section~\ref{ssec:settings} Apparatus, rather than a throughput-optimized local deployment.
In practical deployments, the LLM-based offline indexing stage can be substantially accelerated by batching requests or locally serving the model with high-throughput inference frameworks such as vLLM.

\begin{table}[ht]
\centering
\small
\begin{tabularx}{\columnwidth}{l>{\centering\arraybackslash}X}
\toprule
\textbf{Metric} & \textbf{Value per SQL} \\
\midrule
\multicolumn{2}{c}{\textbf{Offline Indexing}} \\
SQL Parsing Time & 2.137 ms \\
Description Generation Time & 5181 ms \\
Embedding Time & 3.048 ms \\
Storage & 20.62 KB \\
\midrule
\multicolumn{2}{c}{\textbf{Online Retrieval}} \\
Keyword Extraction Time & 1918 ms \\
Retrieval Time & 53.96 ms \\
Filter and Rerank Time & 4062 ms \\
\bottomrule
\end{tabularx}
\caption{Efficiency analysis of our proposed retriever. Since our retriever consists of multiple stages, we report the efficiency of each step separately. The offline indexing phase is performed only once during knowledge base construction and is not repeated for each text-to-SQL task, so we report it separately from the online retrieval phase.}
\label{tab:efficiency_analysis}
\end{table}

\subsection{Case Studies}
\label{appssec:case_studies}

To address concerns about whether historical knowledge genuinely helps and how retrieval noise impacts performance, we present two representative cases with contrasting outcomes. Both cases achieved 100\% recall, yet produced different results due to retrieval precision and LLM behavior.

\textbf{Success Case (question\_id=1145, ``student\_club'' database)}: Identify the full name and hometown (city, county, state) of the Secretary who attended a social event held at ``900 E. Washington St.'' and also generated the highest total income for the club.

Ground-truth SQL (20 lines omitted):

\begin{lstlisting}[language=SQL]
WITH MemberIncome AS (
    SELECT T1.first_name,
        T1.last_name,
        T1.zip...
),
...
SELECT MI.first_name,
    MI.last_name,
    ZC.city,
    ZC.county,
    ZC.state
FROM MemberIncome AS MI ...
INNER JOIN zip_code AS ZC
ON MI.zip = ZC.zip_code
\end{lstlisting}

Precision / Recall / F1: 90.0\% / 100\% / 94.7\%.

Key Retrieval Results (1 of 10 shown): ``\textit{Find the city and state of the club president based on their zip code}'' $\rightarrow$ \texttt{SELECT ... FROM member AS T1 INNER JOIN zip\_code AS T2 ON T1.zip = T2.zip\_code...} (question\_id=1333 in the original BIRD-Dev dataset)

Why This Succeeded: The key retrieved result provided the \textbf{non-obvious foreign key relationship} \texttt{member.zip} $\rightarrow$ \texttt{zip\_code.zip\_code}, which an LLM cannot infer from column names alone.
Without this historical signal, the hometown lookup would fail.

\textbf{Failure Case (question\_id=1213, ``superhero'' database)}:
List the names, heights, and eye colors of male Marvel Comics superheroes who have the maximum strength attribute value, ranked by height from tallest to shortest.

Ground-truth SQL (23 lines omitted):

\begin{lstlisting}[language=SQL]
... SELECT ha.hero_id
FROM hero_attribute AS ha INNER JOIN ...
WHERE ... AND ha.attribute_value = (
    SELECT MAX(ha2.attribute_value)
    FROM hero_attribute AS ha2
    INNER JOIN ...
) ...
\end{lstlisting}

Generated SQL (13 lines omitted):

\begin{lstlisting}[language=SQL]
SELECT ... FROM superhero AS T1
INNER JOIN hero_attribute AS T5
ON T1.id = T5.hero_id ...
ORDER BY T5.attribute_value DESC
LIMIT 1
\end{lstlisting}

Precision / Recall / F1: 28.6\% / 100\% / 44.4\%

Key Retrieval Results (2 of 14 shown):

\begin{itemize}[itemsep=0pt, parsep=0pt]
\item ``\textit{Find the full name of the superhero with the \textbf{highest strength attribute} value}'' $\rightarrow$ \texttt{SELECT ... ORDER BY attribute\_value DESC LIMIT 1} (\textbf{False positive})
\item ``\textit{Count the number of superheroes who have the \textbf{highest strength attribute} value}'' $\rightarrow$ \texttt{SELECT ... WHERE attribute\_value = (SELECT MAX(attribute\_value) FROM hero\_attribute)} (\textbf{True positive})
\end{itemize}

Why This Failed: Despite 100\% recall, the LLM selected the wrong knowledge item when multiple valid options were present in the retrieval context. 
Specifically, both \texttt{attribute\_value = (SELECT MAX(...))} and the oversimplified \texttt{ORDER BY attribute\_value DESC LIMIT 1} appeared in the retrieved examples.
The LLM erroneously chose the latter, which returns exactly one result instead of all heroes tied at the maximum strength value.
Hence, improving retrieval precision is critical for history-driven Text-to-SQL.

\section{Prompt Templates}
\subsection{Prompt Templates for Dataset Construction}

This section presents the prompt templates used in the dataset construction pipeline (Section~\ref{sssec:auto-task-gen}).
The pipeline consists of three main components: Task generator, knowledge validator, and task fixer.

\subsubsection{Task Generator}

The task generator is responsible for creating SQL tasks by selecting and reusing components from a set of original SQL queries.

\begin{lstlisting}[
caption={Task generation instruction.},
label={lst:task_generator},
basicstyle=\ttfamily\small,
breaklines=true,
frame=single,
backgroundcolor=\color{gray!10},
keywordstyle={},
commentstyle={},
stringstyle={},
identifierstyle={},
numberstyle={},
ndkeywordstyle={},
showspaces=false,
showstringspaces=false]
You are given a set of **SQL script**, their **Natural language explanation**, their **Evidence**, their **components**, and **Schemas**. Your task is to reuse part of the **SQL script** to create a SQL task. The task consists of a task description and a SQL query that matches the description. You should think step by step when creating the task:
1. Randomly select 2-6 provided SQLs (i.e., the length of "used_ori_sqls" list below should be 2-6). In order to ensure that the generated SQL contains rich domain knowledge, you should try to choose SQL statements with dissimilar meanings.
2. randomly select interesting SQL components from the SQLs selected in step 1, such as complex joins, nested queries, aggregations, compound conditions, or even the whole SQL query.
3. Randomly tweak some of the selected components (e.g., modify the literal values) according to the **Schemas**. This is because the task you create is not simply the combination of components, but a new SQL query that reuses the knowledge/logic behind these components.
4. Think of a task description that can be solved using some of these components. The task SQL (about 30 lines) should be harder than the original SQLs (typically 7-15 lines). The description should be concise and natural, but cannot be too broad or too specific. You can imitate the style of the given **Natural language explanation** (which means you do not need to explain everything in the task description, and leave some details in the task evidence).
5. Reuse the components (some of them are tweaked) to create new SQL queries (about 30 lines) that exactly match the task descriptions.

Strict JOIN constraints for task creation:
- Do not introduce more joined tables than any single original you reuse. For example, if originals are 'A JOIN B JOIN C' and 'A JOIN C JOIN D', producing 'A JOIN B JOIN C JOIN D' is invalid. Subsets like 'A JOIN C' are allowed.
- Do not change JOIN modes arbitrarily (e.g., INNER vs LEFT) unless this change is clearly implied by the originals' semantics.
- Avoid chains of four or more consecutive JOIN clauses; instead, restructure the query with CTEs or subqueries.

Schemas:
## Schemas

Table: <table_name>
Column description:
<column_description>

Table: <table_name>
Column description:
<column_description>
...

Original SQLs:
## SQL script

### SQL 0

Natural language explanation: <description>
Evidence: <evidence>
Script:

```sql
<formatted_sql>
```

Components:

```json
[
    {
        "idx": <int>,
        "clause": "<clause_name>",
        "type": "<type>",
        "sql_substring": "<sql_substring>"
    },
    ...
]
```

### SQL 1
...

Output in JSON format:
{
    "task": {
        "used_ori_sqls": <List[int]. List of indices of the original SQLs used.>,
        "description": "<Str. Task description>",
        "evidence": "<Str. Task evidence>",
        "sql": "<Str. SQL query>"
    }
}
\end{lstlisting}

\subsubsection{Knowledge Validator}

The knowledge validator checks whether the generated task meets quality standards and meaningfully reuses knowledge from the original SQLs.

\begin{lstlisting}[
caption={Validation prompt template.},
label={lst:knowledge_validator},
basicstyle=\ttfamily\small,
breaklines=true,
frame=single,
backgroundcolor=\color{gray!10},
keywordstyle={},
commentstyle={},
stringstyle={},
identifierstyle={},
numberstyle={},
ndkeywordstyle={},
showstringspaces=false]
You are a meticulous SQL task validator.

Goal:
- Check if the candidate task passes the quality checklist.

Original SQLs:
### Original SQL 0
NL: <natural_language_description>
Evidence: <evidence>
Script:
```sql
<formatted_sql>
```

### Original SQL 1
...

Candidate:
### Candidate Task
Description: <task_description>
Evidence: <task_evidence>
SQL:
```sql
<formatted_task_sql>
```

Execution Feedback:
<execution_feedback>

Automatic diagnostics:
- Execution success: Yes/No
- Non-empty result: Yes/No
- Any one-line stats with 0/NA: Yes/No
- All rows 0/NA: Yes/No
- SQL length out of bounds: Yes/No
- Max consecutive JOIN clauses: <number>
- Exceeds JOIN chain limit: Yes/No

Checklist:
- Must execute successfully (syntax-valid and runnable).
- Must return a non-empty result set.
- Results must be reasonable (not all zeros/NA; no degenerate values).
- Every clause must be purposeful; you should be able to explain each part.
- JOIN constraints: do not introduce more joins/tables than any original; subsets are allowed. Example: originals have 'A JOIN B JOIN C' and 'A JOIN C JOIN D': producing 'A JOIN B JOIN C JOIN D' is invalid (more tables), while 'A JOIN C' is allowed (subset).
- Do not change JOIN modes arbitrarily (e.g., INNER vs LEFT) unless clearly implied by the originals.
- Do not introduce function categories absent in the originals (e.g., originals only use aggregations: avoid string/window functions).
- Preserve CASE WHEN semantics; especially keep the intent of THEN expressions (e.g., do not change THEN molecule_id to THEN 1).
- Do not rewrite COUNT(DISTINCT ...) as COUNT(...) or other forms.
- WHERE/HAVING filters may be recombined from originals; value changes are acceptable if consistent with domain logic.
- GROUP BY should accompany aggregations; it may be implicit even if not mentioned in NL.
- ORDER BY: keep only if (a) explicitly required by NL, or (b) paired with LIMIT for min/max or top/bottom patterns; otherwise remove.
- Avoid obvious inefficiencies (redundant joins, unnecessary subqueries, duplicate work).
- Difficulty: target ~20 lines; hard cap at 30 lines.
- Critically: the new SQL must meaningfully reuse implicit knowledge from the originals (domain logic, computation patterns), not superficial tokens.

Return only JSON with fields:
{
  "pass": true|false,
  "issues": [
    { "rule": "<short name>", "severity": "minor|major", "explanation": "<why>" }
  ],
  "recommendations": "<short actionable advice>"
}
\end{lstlisting}

\subsubsection{Task Fixer}

The task fixer addresses issues identified during validation and execution, refining the task to meet quality standards.

\begin{lstlisting}[
caption={Task fixing instruction.},
label={lst:task_fixer},
basicstyle=\ttfamily\small,
breaklines=true,
frame=single,
backgroundcolor=\color{gray!10},
keywordstyle={},
commentstyle={},
stringstyle={},
identifierstyle={},
numberstyle={},
ndkeywordstyle={},
showstringspaces=false]
The **SQL task** you created above may contain some issues. Here are the possible issues and their solutions:
a. If there are execution errors, please check the SQL syntax and simply modify the SQL to make sure that it can run successfully.
b. If the SQL can run successfully, but the execution result is empty, please modify the task (including the description, evidence, and SQL) to ensure that the SQL query exactly matches the task description and returns non-empty results. Usually you can remove/change some of the filtering conditions to achieve this.
c. If the SQL returns results, please carefully check whether the results are reasonable according to the task description (for example, if the results are all zero or NA, the task description may not be reasonable). If not reasonable, please modify the task (including the description, evidence, and SQL) to ensure that the SQL query exactly matches the task description.
d. Please check if the SQL query can be optimized since some SQLs have redundant joins/subqueries/CTEs, please optimize them to make the SQL more concise while ensuring that the SQL exactly matches the task description.
e. If there is a hint in **SQL execution feedback** saying that the SQL is too complex or too simple, please consider another revision of the task (including the description, evidence, and SQL) to make the difficulty of the task more appropriate. 

If you are absolutely certain that the current SQL is correct and optimal, you can also choose not to make any changes.

Please also follow the previous step 1-5 and previously given schemas when revising the task.

Strict JOIN constraints for revision:
- Do not introduce more joined tables than any single original you reused; subsets are allowed (e.g., 'A JOIN C').
- Do not change JOIN modes arbitrarily (e.g., INNER vs LEFT) unless clearly implied by the originals.

Output in the same JSON format as before.
\end{lstlisting}

\subsection{Prompt Templates for Our Baseline Approach}

This section presents the prompt templates used in the knowledge extraction and Retrieval pipeline (Section~\ref{sec:baseline-approach}).
The pipeline consists of three main components: Knowledge Extraction, Keyword Extraction, and Filter-and-Rerank.

\subsubsection{Knowledge Extraction}
\label{appsssec:know_ext_prompt}

The Knowledge Extraction component mentioned in Section~\ref{sec:baseline-approach} extracts domain knowledge from SQL scripts by converting code-based concepts into natural language-based concepts.

\begin{lstlisting}[
caption={Knowledge Extraction Instruction},
label={lst:knowledge_extraction},
basicstyle=\ttfamily\small,
breaklines=true,
frame=single,
backgroundcolor=\color{gray!10},
keywordstyle={},
commentstyle={},
stringstyle={},
identifierstyle={},
numberstyle={},
ndkeywordstyle={},
showstringspaces=false]
You are given a **SQL script**, **Schemas** and a series of **Concepts** represented in code snippets in the SQL script. Your task is to convert these code-based Concepts into NL-based Concepts as the domain knowledge of authoring SQL queries. Such NL-based Concepts should be easily understandable by humans and easily searchable by machines.

## Instructions

(Inst. 1) Generate an NL description of the **SQL script** according to the **Schemas** provided. The description should clearly tell the query task but do not include step-by-step data transformation details. Example: "List the team names which have at least 3 all-star players."
(Inst. 2) Given **Concepts** consisting of idx, type, and sql_substring, generate names, aliases, descriptions of each given concept respectively. The names can explicitly or implicitly come from NL description of Inst. 1. Think of 0-3 synonyms of the concept name as aliases so that the concept can be more easily searched. For descriptions, you can refer to the following templates:
1. Relation: Which tables are involved in the query and how they are connected. Usually you can find them in FROM-JOIN-ON statement.
Template: Should [join type] [Names of tables] on [names of columns] to [purpose].
Example: Should inner join players_teams and player_allstar ON playerID to obtain team ID.
2. Calculation: How to calculate a numerical/categorical attribute using arithmetic, aggregation functions or non-aggregation functions. Usually you can find them in SELECT statement.
Template: [name of attribute] = [formula/pseudocode].
Example: percentage of players = Divide(Count(playerID where birthState = 'NY'), Count(playerID)) * 100.
3. Dimension: The granularity of querying. Usually you can find them in GROUP BY statement.
Template: Should consider [names of columns] when aggregating [some attributes].
Example: Should consider both playerID and birthDate when aggregating most MVPs. 
4. Condition: Which filters are applied to the data according to modifiers in the natural language query. Usually you can find them in WHERE/HAVING statement.
Template: "[A phrase or word that serves as a modifier]" refers to [formula/pseudocode].
Example: "exceed 75% of defensive rebounds" refers to Divide (dRebounds, rebounds) * 100 > 75"
5. Output: Specify the rows to be shown, the order, whether to deduplicate, etc. Usually you can find them in ORDER BY/LIMIT/DISTINCT statement.
Template: Consider [some modifiers] records.
Example: Consider only 10 distinct records.

Schemas:
## Schemas

Table: <table_name>
Column description:
<column_description>

Table: <table_name>
Column description:
<column_description>
...

SQL script:
## SQL script

```sql
<formatted_sql>
```

Concepts:
## Concepts

```json
[
    {
        "idx": <int>,
        "type": "<type>",
        "sql_substring": "<sql_substring>"
    },
    ...
]
```

Output in JSON format like (if **Concepts** are empty, simply assign an empty list to "concepts" field):

```json
{
    "nl_description": "<Use one or several imperative sentences to describe the query task>",
    "concepts": [
        {
            "idx": <Int. The index of the concept provided>,
            "name": <Str. An English phrase or word. The phrase should not be too broad (e.g. "output format"). Try to be specific and relevant to the business context (e.g. "field goals"). If there are several words in a phrase please separate them with whitespace>,
            "alias": <List[str]. Also English phrases or words that are the synonyms of the concept name>,
            "description": <Str. Explain the concept using a short NL sentence. You can refer to the template mentioned above>
        },
        ...
    ]
}
```
\end{lstlisting}

\subsubsection{Keyword Extraction}

The Keyword Extraction component extracts meaningful keywords from natural language queries to facilitate candidate retrieval.

\begin{lstlisting}[
caption={Keyword Extraction Instruction},
label={lst:keyword_extraction},
basicstyle=\ttfamily\small,
breaklines=true,
frame=single,
backgroundcolor=\color{gray!10},
keywordstyle={},
commentstyle={},
stringstyle={},
identifierstyle={},
numberstyle={},
ndkeywordstyle={},
showstringspaces=false]
You are an expert SQL writer. Now you need to extract some keywords from the natural language query written by users: "<nl_query>". Focus on extracting meaningful phrases/concepts rather than individual words. For example, you can prioritize these types of phrases/concepts:
- Complete noun phrases
- Important conditions or constraints
- Key relationships between entities
- Specific measures or attributes
      
You should not extract more than 8 keywords, usually 1-5 keywords are enough. Output in JSON format like:

```json
{
    "keywords": <List[str]. The list of keywords extracted from the natural language query. Each keyword should be a single word or a short phrase.>
}
```

## Schemas

Table: <table_name>
Column description:
<column_description>

Table: <table_name>
Column description:
<column_description>
...

## Your output
\end{lstlisting}

\subsubsection{Filter and Rerank}

The Filter-and-Rerank component selects and reorders relevant SQL candidates and concepts based on the natural language query.

\begin{lstlisting}[
caption={Filter-and-Rerank Instruction},
label={lst:filter-rarank},
basicstyle=\ttfamily\small,
breaklines=true,
frame=single,
backgroundcolor=\color{gray!10},
keywordstyle={},
commentstyle={},
stringstyle={},
identifierstyle={},
numberstyle={},
ndkeywordstyle={},
showstringspaces=false]
You are an expert in SQL who wants to reuse some code segments from existing SQL scripts to build new queries. You are given a **Natural language query**, **Schemas** and some **Candidates**. The explanation of each field in the **Candidates** is:
- sql: The SQL script
- filename: The filename of the SQL script
- nl_description: The natural language description of the SQL script
- concepts: A list of concepts. Each concept has an index, a name, a type ("relation", "calculation", "condition", "dimension", "output"), and a SQL substring to which the concept refers. The concepts serve as domain knowledge that is important for understanding/writing the SQL script, so even if you do not use the whole SQL script, you can still use some of the concepts.

You need to select relevant SQL scripts and/or relevant concepts from the following **Candidates**. Output in JSON format. The items of "filtered_candidates" below should be reordered by the relevance (put the most relevant candidates first). If there are no relevant candidates, you should make "filtered_candidates" an empty list. You should try to avoid selecting semantically duplicate concepts (E.g. If query A has the concept A.X, A.Y, and query B has the concept B.Y, B.Z, and you only want to reuse the concept Y. In that case, you should only select either A.Y or B.Y, but not both).

## Output Format

```json
{
    "filtered_candidates": [
        {
            "filename": <String. The filename of the SQL candidate>,
            "use_all": <Bool. Whether to use the entire SQL script>,
            "indices": <List[int]. The list of concept indices that are relevant to the query. You can refer to the `idx` attribute of the each concept item. If use_all is True, this list should be empty.>
        },
        {...},
        ...
    ]
}
```

## Natural language query

<nl_query>

## Schemas

Table: <table_name>
Column description:
<column_description>

Table: <table_name>
Column description:
<column_description>
...

## Candidates

```json
[
    {
        "sql": "<sql_script>",
        "filename": "<filename>",
        "nl_description": "<nl_description>",
        "concepts": [
            {
                "idx": <int>,
                "name": "<name>",
                "type": "<type>",
                "sql_substring": "<sql_substring>"
            },
            ...
        ]
    },
    ...
]
```

## Your output
\end{lstlisting}

\end{document}